%% file: 0_main.tex
\documentclass[10pt,twocolumn,letterpaper]{article}

\usepackage{iccv}
\usepackage{times}
\usepackage{epsfig}
\usepackage{graphicx}
\usepackage{amsmath}
\usepackage{amssymb}

\newcommand\blfootnote[1]{%
  \begingroup
  \renewcommand\thefootnote{}\footnote{#1}%
  \addtocounter{footnote}{-1}%
  \endgroup
}

\usepackage{xcolor}
\usepackage{makecell}
\usepackage{enumitem}
\usepackage{gensymb}
\usepackage{multirow}
\usepackage{amsmath,amssymb}
\usepackage{wrapfig, lipsum}
\usepackage{ragged2e}
\usepackage{pifont}%
\usepackage{xspace}
\usepackage{csquotes}

\def\ours{\texttt{\textbf{DIME-FM}}\xspace}
\def\smallclip{Distill-ViT-B/32\xspace}

\definecolor{Green}{rgb}{0.0, 0.5, 0.0}
\definecolor{darkcyan}{HTML}{76A5AF}
\definecolor{peru}{HTML}{E69138}
\definecolor{indigo}{HTML}{674EA7}
\definecolor{blue1}{HTML}{6D9EEB}
\definecolor{Red}{rgb}{1, 0.6, 0.6}
\definecolor{Blue}{rgb}{0, 0.6, 1}
\definecolor{PaleYellow}{rgb}{0.8, 0.8, 0}

\makeatletter
\let\original@footnote\footnote
\newcommand{\align@footnote}[1]{%
  \ifmeasuring@
    \chardef\@tempfn=\value{footnote}%
    \footnotemark
    \setcounter{footnote}{\@tempfn}%
  \else
    \iffirstchoice@
      \original@footnote{#1}%
    \fi
  \fi}
\pretocmd{\start@align}{\let\footnote\align@footnote}{}{}
\makeatother

\usepackage[pagebackref,breaklinks,colorlinks]{hyperref}

\usepackage[capitalize]{cleveref}
\crefname{section}{Sec.}{Secs.}
\Crefname{section}{Section}{Sections}
\Crefname{table}{Table}{Tables}
\crefname{table}{Tab.}{Tabs.}

\iccvfinalcopy %
\def\iccvPaperID{3628} %

\ificcvfinal\fi

\input{mlVecMat}

\begin{document}

\title{DIME-FM : DIstilling Multimodal and Efficient Foundation Models}
\author{Ximeng Sun$^{1\dagger}$ \ \ \ \  Pengchuan Zhang$^{2}$  \ \ \ \ Peizhao Zhang$^{2}$ \ \ \ \  Hardik Shah$^{2}$ \ \ \ \  Kate Saenko$^{1,2}$ \ \ \ \ Xide Xia$^{2}$\\
$^{1}$ Boston University, $^{2}$ Meta AI \\
}
\maketitle
\ificcvfinal\thispagestyle{empty}\fi

\begin{abstract}
   Large \textbf{V}ision-\textbf{L}anguage \textbf{F}oundation \textbf{M}odels (VLFM), such as CLIP, ALIGN and Florence, are trained on large-scale datasets of image-caption pairs and achieve superior transferability and robustness on downstream tasks, but they are difficult to use in many practical applications due to their large size, high latency and fixed architectures.  \blfootnote{$^\dagger$Work done when interning at Meta AI.}  Unfortunately, recent work shows training a small custom VLFM for resource-limited applications is currently very difficult using public and smaller-scale data. In this paper, we introduce a new distillation mechanism (\ours) that allows us to transfer the knowledge contained in large VLFMs to smaller, customized foundation models using a relatively small amount of inexpensive, unpaired images and sentences.
   \textbf{We transfer the knowledge from the pre-trained CLIP-ViT-L/14 model to a ViT-B/32 model, with only 40M public images and 28.4M unpaired public sentences}. The resulting model ``\smallclip'' rivals the CLIP-ViT-B/32 model pre-trained on its private WiT dataset (400M image-text pairs): 
    \smallclip achieves similar results in terms of zero-shot and linear-probing performance on both ImageNet and the ELEVATER  (20 image classification tasks) benchmarks. It also displays comparable robustness when evaluated on five datasets with natural distribution shifts from ImageNet. Please refer to our \href{https://cs-people.bu.edu/sunxm/DIME-FM/}{project page} for code and more details. 
\end{abstract}

\input{1_introduction}

\input{2_related_works}

\input{3_method}

\input{4_experiments}

\input{5_conclusion}

\clearpage

\newpage
\clearpage

{\small

\input{dime_fm.bbl}
}

\clearpage
\newpage
\appendix

\begin{table} [h]
\centering 
\begin{adjustbox}{max width=\linewidth}
\begin{tabular}{ c|c }
 \toprule
Section & Content    \\
\midrule
\ref{sec:implementation_details} & Implementation Details \\
\ref{sec:visualization} & Visualization of the Constructed $\Tcal$ \\
\ref{sec:contribution_nlp} & Contribution of each NLP dataset \\
\ref{sec:dataloading} & Paired vs. Unpaired Dataloading  \\
\ref{sec:concept_coverage} & Conceptual Coverage Analysis\\
\ref{sec:mmd} &  MMD among Image and Text Corpora \\
\ref{sec:full_ablation_on_losses} &  Full Ablation Studies on Losses \\
\ref{sec:detailed_performance} & Detailed Performance on Each Dataset\\
\bottomrule
\end{tabular}
\label{tab:supp}
\end{adjustbox}
\caption{\textbf{Supplementary Material Overview}}
\vspace{-15pt}
\end{table}

\input{A1_implementation_details.tex}

\input{A3_visualization_image_text.tex}

\input{A4_contribution_NLP.tex}

\input{A5_paired_vs_unpaired_dataloading.tex}

\input{A6_analysis_of_elevator.tex}

\input{A2_mmd_score_among_corpora.tex}

\input{A8_full_ablation_study.tex}

\input{A7_detailed_information.tex}

\end{document}

%% file: mlVecMat.tex
\usepackage{amsmath}
\usepackage{amsfonts}
\usepackage{amssymb}
\usepackage{wrapfig}
\usepackage{subcaption}
\usepackage{multirow}
 \usepackage{mathtools} 

\usepackage{verbatim}

\usepackage{anyfontsize}

\usepackage{microtype}
\usepackage{graphicx}
\usepackage{booktabs} %
\usepackage{multirow}
\usepackage{amsmath,amssymb}
\usepackage{booktabs}
\usepackage{caption,subcaption}
\usepackage{xcolor}
\usepackage{enumitem}
\setitemize{itemsep=10pt,topsep=0pt,parsep=0pt,partopsep=0pt}
\usepackage{adjustbox}

\newcommand{\RN}[1]{%
	\textup{\lowercase\expandafter{\it \romannumeral#1}}%
}
\usepackage{tabu}

\newcommand{\beq}{\vspace{0mm}\begin{equation}}
\newcommand{\eeq}{\vspace{0mm}\end{equation}}
\newcommand{\beqs}{\vspace{0mm}\begin{eqnarray}}
\newcommand{\eeqs}{\vspace{0mm}\end{eqnarray}}
\newcommand{\barr}{\begin{array}}
\newcommand{\earr}{\end{array}}

\newcommand{\tv}[0]{{\boldsymbol{t}}}
\newcommand{\uv}{\boldsymbol{u}}
\newcommand{\vv}{\boldsymbol{v}}
\newcommand{\wv}{\boldsymbol{w}}
\newcommand{\xv}{\boldsymbol{x}}

\newcommand{\thetav}{\boldsymbol{\theta}}

\newcommand{\phiv}{\boldsymbol{\phi}}

\newcommand{\R}{\mathbb{R}}

\newcommand{\Xcal}{\mathcal{X}}

\newcommand{\Tcal}{\mathcal{T}}

\newcommand{\Lcal}{\mathcal{L}}

\newcommand{\Vcal}{\mathcal{V}}

\newcommand{\Ucal}{\mathcal{U}}

\usepackage{color, colortbl}
\definecolor{Gray}{gray}{0.93}

\usepackage{xcolor,amsmath}
\usepackage[linesnumbered,ruled,vlined]{algorithm2e}
\SetKwRepeat{Do}{do}{while}%
\DontPrintSemicolon

\SetKwComment{Comment}{\color{green!50!black}\# }{}

\SetKwProg{Function}{def}{:}{}

\SetKwProg{For}{for}{:}{}
\SetKwProg{If}{if}{:}{}

\DeclareMathOperator*{\argmax}{arg\,max}

%% file: 1_introduction.tex
\section{Introduction}
In contrast to neural networks learnt to solve a single target vision task (\ie task-specific models)~\cite{he2016deep, long2015fully, tan2019efficientnet, chen2022systematic, Sun_2021_ICCV, liu2016ssd},  CLIP~\cite{radford2021learning} and other \textbf{V}ision-\textbf{L}anguage ``\textbf{F}oundation \textbf{M}odels''~(VLFMs)~\cite{li2021align,yuan2021florence} achieve superior accuracy on diverse novel downstream tasks and improved robustness to natural domain shifts during inference. At the same time, small and customizable VLFMs are in high demand for many applications that have limited computational resources (AV, AR/VR and other edge devices).  Unfortunately, only a few labs in the world can afford the  large-scale vision-language  datasets (e.g. WiT~\cite{radford2021learning} with 400M image-text pairs) and the immense computing resources required to train VLFMs.  Efforts to re-create VLFMs on public data {\cite{yang2022unified, shen2022klite, cherti2022reproducible}) either fall short on accuracy or require even more expensive training on huge datasets of images paired with captions (\eg over 5B pairs~\cite{schuhmann2022laion}).

\begin{figure}[t]
\begin{center}
     \includegraphics[width=1\linewidth]{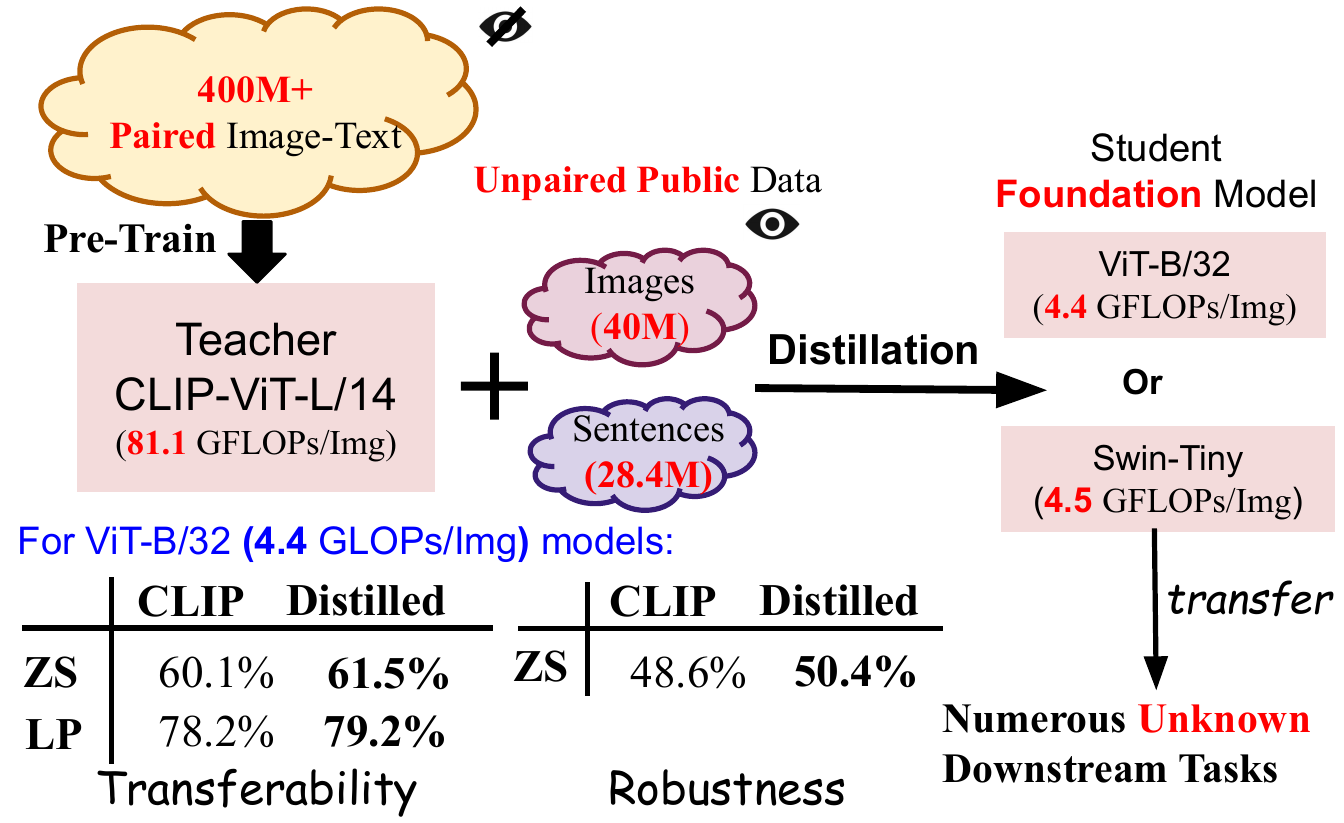}
\end{center} 
\vspace{-15pt}
   \caption{\small 
   \textbf{Conceptual Figure of our Vision-Language Knowledge Distillation  \ours .} We distill the knowledge from a large VLFM ``CLIP-ViT-L/14' pretrained on 400M private image-text paired dataset.  We only use public unpaired image and text corpora as inputs. Our Distill-ViT-B/32 rivals CLIP-ViT-B/32 in both transferability and robustness. ZS: Zero-Shot, LP: Linear Probing. 
   }
   \label{fig:concept_figure.}
   \vspace{-10pt}
\end{figure}

Instead of pretraining, model distillation used to offer a convenient way to obtain a smaller custom model. Recent work distills CLIP specifically for one or a few target tasks (\ie task-specific distillation).  For example, 
some works~\cite{wei2022mvp,wei2022contrastive,zhang2022cae,peng2022beit} distill the CLIP's image feature maps for better visual feature representations. 
BeamCLIP~\cite{kimtransferring} distills  CLIP logits for a single target image classification task, \eg ImageNet-1K~\cite{imagenet_cvpr09}. More recently, CLIP-TD~\cite{wang2022clip} distills CLIP to solve three specific vision-language tasks. 
Even though these task-specific distillation works achieve good performance for the specialized downstream task, they are not scalable to solve new downstream tasks by zero-shot transferring. 
\textbf{There is no approach for distilling VLFMs to another foundation model} which preserves transferability to \textit{novel} tasks and robustness to domain shifts. Due to the unaffordable large-scale pretraining and the lack of foundation-model distillation mechanism, \textbf{practitioners must rely on the few labs to release smaller VLFMs, and cannot easily customize their size or architecture.}

In this work, we successfully distill smaller custom VLFMs using only smaller-scale public data, but achieving comparable transferability and robustness as if they were pre-trained on the large-scale data.
Specifically , we transfer the knowledge from the released CLIP-ViT-L/14~\cite{radford2021learning} to our small VLFM \enquote{Distill-ViT-B/32}. During the distillation, we adopt only \textbf{40M images from public datasets and  28.4M unpaired sentences}.
Remarkably, with less than one-tenth of CLIP's pretraining dataset WiT, \smallclip achieves comparable transferability and robustness to CLIP-ViT-B/32~\cite{radford2021learning} (see Fig.~\ref{fig:concept_figure.}).

 To accomplish this, we propose a novel distillation mechanism to \textbf{DI}still \textbf{M}ultimodal and \textbf{E}fficient \textbf{F}oundation \textbf{M}odels (\ours) from CLIP. In standard distillation of 
image classification models with the fixed categories (\ie fixed-vocabulary models),  the class scores (logits) are matched between the teacher and student models~\cite{hinton2015distilling,kim2018paraphrasing,ba2014deep,mirzadeh2020improved,beyer2022knowledge}. However,  since VLFMs do not have fixed-vocabulary logits, we instead match similarity of images to sentences (\ie open-vocabulary logits)
to retain the transferability (especially zero-shot ability) and robustness of VLFMs.  
 We perform a careful ablation study of how \enquote{vocabulary}, determined by training sentences, affects the student model’s performance and find that it is crucial to perform distillation with a visually-related vocabulary rather than a random vocabulary. To construct a visually-related distillation text corpus, we propose an efficient algorithm that selects visually-grounded sentences (\ie sentences which describe the visual world) from an NLP corpus rather than require the expensive human-annotated image
captions or use noisy web-crawled image-related text.  On top of text selection algorithm, we design two distillation losses to augment open-vocabulary logits in VLFM and empirically show that our novel distillation losses benefit vision-language (VL) knowledge distillation.

To summarize, we make three contributions in this paper: 
\begin{enumerate}[leftmargin=*]
    \item  We propose a vision-language knowledge distillation mechanism \ours to transfer knowledge of pre-trained huge VLFMs to \textbf{small  foundation models} with smaller-scale public images and unpaired sentences. 
    \item We distill the pre-trained CLIP-ViT-L/14 to \smallclip, with only unpaired 40M public images and 28.4M sentences. Notably, our \smallclip rivals the CLIP-ViT-B/32 that was pre-trained on private 400M image-text paired data in both transferability and robustness.
     \item Our proposed \ours consists of an efficient algorithm to construct a visually-grounded text corpus from an NLP corpus and two specific distillation losses to augment open-vocabulary logits in VL distillation.

\end{enumerate}

%% file: 2_related_works.tex
\section{Related Works}
\vspace{-1mm}
\noindent \textbf{Vision-Language Foundation Models.} Many previous works focus on learning a generic alignment between language and vision features extracted by pretrained encoders~\cite{gordo2017beyond, kim2021vilt, li2020oscar, lu2019vilbert,su2019vl, wang2021simvlm,zhang2021vinvl} to improve many downstream tasks, \eg Visual Question Answering (VQA)~\cite{antol2015vqa, xu2016ask, hudson2019gqa}, Image Captioning~\cite{agrawal2019nocaps,lin2014microsoft,rohrbach2018object,hendricks2018women} \textit{etc}. Recently, inspired by the great success on generic NLP model transferring to the downstream tasks~\cite{radford2018improving,radford2019language,brown2020language}, CLIP~\cite{radford2021learning} and other large VLFMs~\cite{jia2021scaling, li2021align, yuan2021florence, li2022scaling, yao2021filip} pretrain on hundreds of million image-text pairs to learn transferable visual representation from natural language supervision with contrastive learning. These works have shown astonishing transferring performance, such as zero-shot and linear probing evaluations, on various downstream tasks~\cite{li2022elevater} as well as a great robustness to the distribution shift from ImageNet~\cite{radford2021learning}.  Without the use of private large-scale data, it is challenging to learn small custom foundation models that possess comparable transferability and robustness. ELEVATER evaluation~\cite{li2022elevater} shows that training VLFMs~\cite{yang2022unified,shen2022klite}  using relative small public datasets ($\le$ 40M image-text pairs) and even with help of external knowledge, \eg WordNet~\cite{miller1998wordnet} and Wiktonary~\cite{meyer2012wiktionary}, cannot close the performance gap in comparison to CLIP~\cite{radford2021learning} or Florence~\cite{yuan2021florence}. Trained with CLIP-Filtered 400M image-text pairs~\cite{schuhmann2021laion}, OpenCLIP~\cite{cherti2022reproducible} still performs worse than CLIP at each model size because of the possible poorer quality of paired data. In this paper, instead of pretraining the model using contrastive loss with paired data, we distill from CLIP-ViT-L/14 to different  models with smaller-scale 
public images and unpaired sentences.

\vspace{1mm}
\noindent \textbf{Uni-modal Knowledge Distillation.} In general, knowledge distillation~\cite{hinton2015distilling} transfers knowledge from one model (teacher) to another (student). It optimizes a student model to match some certain output of the teacher model. With a single modality, there are two main ways of distillations: (1) knowledge distillation of the fixed-vocabulary prediction logits~\cite{hinton2015distilling,kim2018paraphrasing,ba2014deep,mirzadeh2020improved,beyer2022knowledge}. (2). feature distillation on the final or intermediate activation of the network~\cite{romero2014fitnets,huang2017like,ahn2019variational,heo2019comprehensive,zagoruyko2016paying,sun2021dynamic}. In this paper, we do not require the same feature dimension in both teacher and student foundation models. To avoid complex tricks to circumvent the mismatch of feature dimensions using feature distillation methods, we adopt the simple logit distillation for the vision-language distillation. Instead of applying KL divergence loss to fixed-vocabulary logits in the uni-modal logit distillation, we apply KL divergence loss to feature similarity scores (\ie open-vocabulary logits) in VLFMs. Moreover, we still use the uni-modal logit distillation as a regularizer in the distillation.

\vspace{1mm}
\noindent \textbf{Model Distillation from CLIP.} 
Some works~\cite{wei2022mvp,wei2022contrastive,zhang2022cae,peng2022beit} perform feature distillation of CLIP image encoder with Masked Image Modeling~\cite{bao2021beit,dong2021peco,he2022masked,dong2022bootstrapped,baevski2022data2vec,dong2022maskclip,wang2022bevt,xie2022simmim} to learn a new image encoder which claim superior finetuning performance on ImageNet-1K~\cite{imagenet_cvpr09} and ADE20K~\cite{zhou2019semantic}. They ignore the language encoder during the distillation and do not maintain the alignment of image and text in the feature space. BeamCLIP~\cite{kimtransferring} distills the CLIP using logits computed by images from the public image datasets and class names of ImageNet-1K, and achieves better ImageNet-1K Top-1 linear probe accuracy than vision-only self-supervised learning (SSL) methods~\cite{chen2020simple, caron2020unsupervised}. CLIP-TD~\cite{wang2022clip}  distills knowledge from CLIP into existing architectures to solve targeted vision-language
(VL) tasks.  Even though these works achieve better performance in their specific tasks, their student models lose the capability of VLFMs, as they are not scalable to solve new tasks by zero-shot transferring. Instead of distilling CLIP and tuning it for specific downstream task(s), we wish to distill another foundation model from CLIP, and our result model yields the comparable transferability and robustness performance to the foundation models with the similar model size but pretrained on hundreds of million image-text pairs.

%% file: 3_method.tex
\section{Vision-Language Knowledge Distillation}~\label{sec:method}
In this paper, we propose our VL knowledge distillation \ours which uses the public unpaired images and text to distill a small VLFM from a pretrained large VLFM (CLIP-ViT-L/14). First, we mathematically define VLFMs and our VL distillation setting. Then, we introduce our novel training losses and our text construction algorithm.

\vspace{1mm}
\noindent \textbf{Preliminaries.} 
A dual-encoder VLFM consists of an image and text encoder to extract image/text embeddings respectively, then project the image and text embeddings to the common feature space. 
To get  more flexible design choices for the dimensions of the separate image/text feature spaces and the final shared feature space, we separate the image and text projection layers from the image and text feature encoders. Therefore, 
a standard dual-encoder VLFM can be defined as a quartet $[f_\theta, g_\phi, \mathbf{A}, \mathbf{B}]$. $f_\theta$ and $g_\phi$  are image and text encoders which encode the image $\xv$ and text $\tv$ into their own feature spaces (as ${\uv}' \in \R^{d^v}$  and ${\vv}' \in \R^{d^l}$)\footnote{The upper scripts $v$ and $l$ are short for vision and language, respectively.} respectively. $\mathbf{A} \in \R^{d \times d^v}$  and $\mathbf{B} \in \R^{d \times d^l}$ are two linear layers projecting image and text embeddings (${\uv}'$ and ${\vv}'$) to $\uv$ and $\vv$ in  a shared $d$-dim feature space:
\begin{align}
    &{\uv}' = f_{\thetav}(\xv),  \quad  {\vv}' = g_{\phiv}(\tv), \quad \uv = \mathbf{A}{\uv}', \quad \vv = \mathbf{B}{\vv}' 
\end{align}
The similarity score between the image and text embeddings 
\begin{align}
s(\uv, \vv)= \uv^T \vv / (\|\uv\| \|\vv\|) ~\label{eq:similarity_score}
\end{align}
reveals the semantic relationship between image and text encoded in VLFMs. It plays an important role in transferring to downstream tasks and being robust to domain shift. 

\vspace{1mm}
\noindent \textbf{Problem Definition.} 
Given a public unpaired image corpus $\Xcal$ and text corpus $\Tcal$, we distill a small VLFM $[f_{\widehat{\thetav}}, g_{\widehat{\phiv}}, \widehat{\mathbf{A}}, \widehat{\mathbf{B}}]$ from a pretrained large VLFM $[f_\theta, g_\phi, \mathbf{A}, \mathbf{B}]$, where
\begin{align}
    & \widehat{{\uv}}' = f_{\widehat{\thetav}}(\xv)  \in  \R^{\widehat{d}^v},  \quad  \widehat{{\vv}}' = g_{\widehat{\phiv}}(\tv)  \in \R^{\widehat{d}^l}, \\
    & \widehat{\uv} = \widehat{\mathbf{A}}\widehat{{\uv}}' \in \R^{\widehat{d}} , \quad \widehat{\vv} = \widehat{\mathbf{B}}\widehat{{\vv}}' \in \R^{\widehat{d}}, ~\label{eqn:student_projection}
\end{align} 
where $\widehat{(\cdot)}$ is the component in the student model corresponding to $(\cdot)$ in the teacher model. Notably, we can freely choose the image, text and projected embeddings' dimensions ($\widehat{d}^v$, $\widehat{d}^l$ and $\widehat{d}$) in the student VLFM, which can be different from those (${d}^v$, ${d}^l$ and ${d}$) in the teacher VLFM.

In contrast to the expensive pretraining VLFMs~\cite{radford2021learning, li2021align, cherti2022reproducible} with large-scale image-text pairs, we do not require any paired data for optimization.
During the distillation, we match the similarity scores of feature embeddings between the teacher and student VLFMs, which ensures our distilled small image encoder $f_{\widehat{\thetav}}$ is still superior in transferability and robustness, as if it were trained on large-scale paired data. To this end, we propose our VL distillation mechanism \ours including two novel distillation losses (Sec.~\ref{sec:losses}) and an efficient text selection algorithm to construct the training text corpus (Sec.~\ref{sec:choose_text}).

The capacity of CLIP text encoder has few effects on CLIP performance~\cite{radford2021learning}. Also, the inference latency for close-vocabulary downstream visual tasks~\cite{li2022elevater, sun2022dualcoop, hu2023dualcoop++}. 
To make the presentation simple while keeping the essential idea,  we fix the text encoder, i.e., $g_{\widehat{\phiv}} = g_{\phiv}$, and focus on VL knowledge to distill a small custom image encoder $f_{\widehat{\thetav}}$ as a transferable and robust vision backbone. If a small text encoder $g_{\widehat{\phiv}}$ is desired for open-vocabulary downstream tasks, we can apply the proposed method to distill $g_{{\phiv}}$ while fixing $f_{\widehat{\thetav}}$, which is left as an interesting area of future work.

\begin{figure}
\begin{center}
     \includegraphics[width=1\linewidth]{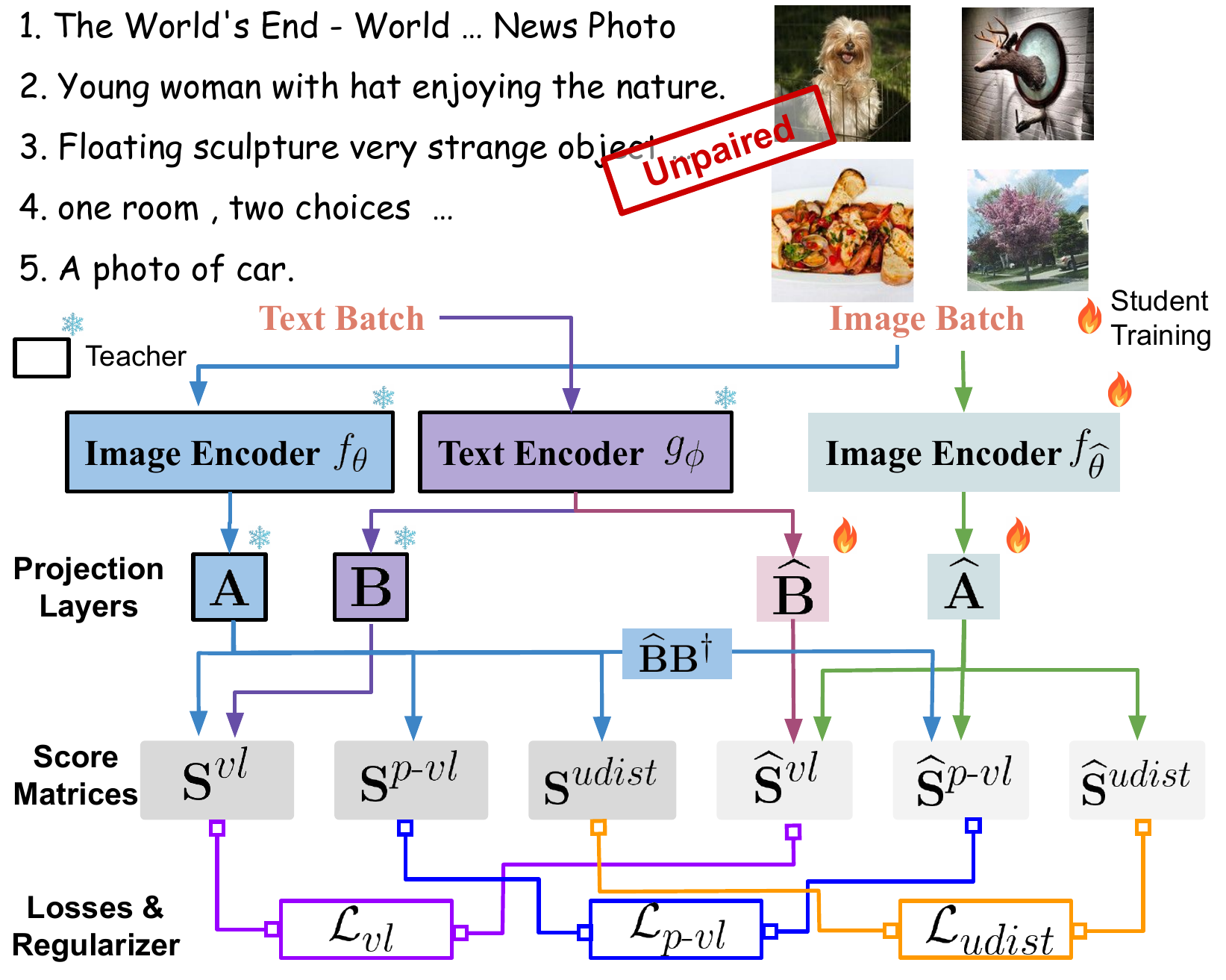}
\end{center} \vspace{-20pt}
   \caption{\small 
   \textbf{ Illustration of our proposed distillation losses.} In each iteration, we compute two losses ($\Lcal_{vl}$, $\Lcal_{p\text{-}vl}$) and one regularizer ($\Lcal_{udist}$) with a min-batch of images and texts to distill knowledge from the teacher to the student. We freeze all parameters in the teacher model and learn the student model from scratch. 
   }
   \label{fig:three_loss}
   \vspace{-10pt}
\end{figure}

\subsection{Optimization for VL Knowledge Distillation}~\label{sec:losses}
In standard uni-modal logit distillation, the objective is to match the fixed-vocabulary logits predicted by the student to the logits predicted by the teacher on the same input sample~\cite{hinton2015distilling}. In VLFMs,  the vocabulary is not fixed and the outputs are similarity score between images and sentences. Thus we change our objective to match the distribution of these scores produced by the student to the distribution produced by the teacher. Specifically, we minimize the KL divergence of score distributions computed over image dataset $\Xcal$ and text dataset $\Tcal$ (see Eq.~\ref{eq:similarity_score}) using three separate losses.

We first define the general form of applying KL divergence to distill the similarity scores and then define the three losses.. Suppose we have two batches of embeddings $\{\wv^1_i\}_{i=1}^{B_1}$ and  $\{\wv^2_j\}_{j=1}^{B_2}$\footnote{ they can be image or text's embeddings. This will be further explained} in the teacher model's shared $d$-dim feature space. All similarity scores form a teacher score matrix $\mathbf{S} \in \R^{B_1 \times B_2}$, where $\mathbf{S} _{i, j} = s(\wv^1_i, \wv^2_j)$.
Similarly, we have the student's score matrix as $\widehat{\mathbf{S} } \in \R^{B_1 \times B_2}$, where $\widehat{\mathbf{S} }_{i, j} = s(\widehat{\wv}^1_i, \widehat{\wv}^2_j)$. Each row and column of the score matrix can be seen as \textit{ open-vocabulary logits}. We measure the row-wise (indexing $i$) and column-wise (indexing $j$)  discrepancy  between $\mathbf{S} $ and $\widehat{\mathbf{S}}$ with KL divergence:
{  
\begin{equation}
\begin{aligned}
\Lcal_{KL}(\widehat{\mathbf{S}}; \mathbf{S}, \mu) & =   \sum\nolimits_{i} \text{KL}( \sigma(\mu \mathbf{S} _i) || \sigma(\mu \widehat{\mathbf{S}_i})) \\
& + \sum\nolimits_{j} \text{KL}(\sigma(\mu \mathbf{S}^T_j) || \sigma(\mu \widehat{\mathbf{S}}^T_j)), \label{eqn:discrepency}
\end{aligned}
\vspace{-2mm}
\end{equation}}
where $\sigma$ is the softmax function and $\mu$ is a temperature.

In particular, we propose two losses ($\Lcal_{vl}$ and  $\Lcal_{p\text{-}vl}$, described in detail below) in form of Eq.~\ref{eqn:discrepency} with different $\mathbf{S}$ and $\mathbf{\widehat{S}}$'s. The third loss is a regularizer $\Lcal_{udist}$ to maintain the Euclidean Distance between every pair of image embeddings (a.k.a geometry of image embeddings) during the distillation. 
Our final VL distillation objective is (see Fig~\ref{fig:three_loss}):
\begin{equation}
    \min\limits_{f_{\widehat{\thetav}}, \widehat{\mathbf{A}}, \widehat{\mathbf{B}}} (1 - \lambda_{1})\Lcal_{vl} + \lambda_{1} \Lcal_{p\text{-}vl} + \lambda_2\Lcal_{udist}~\label{eq:total_loss},
    \vspace{-2mm}
\end{equation}
where $\lambda_1 \in [0,1]$ and $\lambda_2 \in \R^+$ are two hyperparameters to control each loss weight. We study the efficacy of three losses with various $\lambda_1$ and $\lambda_2$'s in Sec.~\ref{sec:ablation_on_losses}.

\vspace{1mm}
\noindent \textbf{VL Score Distillation Loss $\Lcal_{vl}$.} We distill the VL score matrices in form of Eq.~\ref{eqn:discrepency}.  
Given an image batch  $\{\xv_i\}_{i=1}^{B^v} \subset \Xcal$ and a text batch $\{\tv_j\}_{j=1}^{B^l} \subset \Tcal$, they are projected to $\{\uv_i\}_{i=1}^{B^v}$ and $\{\vv_i\}_{i=1}^{B^l}$ in the teacher's shared feature space respectively and projected to $\{\widehat{\uv}_i\}_{i=1}^{B^v}$ and  $\{\widehat{\vv}_i\}_{i=1}^{B^l}$ in the student's feature space. 
Therefore, we define the teacher's and student's VL score matrices as 
\begin{align}
    \mathbf{S}^{vl}_{i, j} = s(\uv_i, \vv_j), \quad \widehat{\mathbf{S}}^{vl}_{i, j} = s(\widehat{\uv}_i, \widehat{\vv}_j),~\label{eqn:mscore}
    \vspace{-2mm}
\end{align}
with which we define VL Score Distillation Loss as:
\begin{align}
    \Lcal_{vl} = \Lcal_{KL}(\widehat{\mathbf{S}}^{vl}, \mathbf{S}^{vl}, \mu^{vl})
\end{align}

\vspace{-1mm}
\noindent \textbf{Pseudo-VL Score Distillation Loss $\Lcal_{p\text{-}vl}$.} 
Our study on the efficacy of text corpus (see Sec~\ref{sec:dataset_scale}) shows that enlarging the text corpus $\Tcal$ introduces more text embeddings and results in more open-vocabulary logits, which in turn benefits the VL knowledge distillation. 

Motivated by this, besides adding more visually-grounded sentences to $\Tcal$, 
we introduce image embeddings as additional pseudo text embeddings. Since image and text embedding are trained to live in a shared sphere (\ie $\forall i, \ \|\uv_i\|_2 = \|\vv_i\|_2 = 1 $), image embeddings are a reasonable substitute for embeddings of visually-grounded text.
For a given image $\xv_j$ and its image embedding $\uv_j$, we assume that there is a sentence $\tv_j$ whose text embedding $\vv_j$ perfectly matches $\uv_j$ in the shared sphere:
\begin{align}
   \vv_j = \uv_j, \quad \vv_j = \mathbf{B}{\vv}_j' \Rightarrow {\vv}_j' \approx \mathbf{B}^{\dagger} \vv_j = \mathbf{B}^{\dagger} \uv_j,~\label{eqn:teacher_pseudo_label}
\end{align}
where $\mathbf{B}^{\dagger}$ is the pseudo-inverse\footnote{also known as Moore–Penrose inverse} of matrix $\mathbf{B}$.
We treat the image embedding $\uv_j$ as the pseudo paired text embedding of the input image $\xv_j$ in the teacher model. For the student model, based on Eq.~\ref{eqn:student_projection}, \ref{eqn:teacher_pseudo_label} and ${\vv}_j' = \widehat{{\vv}'}_j$ (due to the fixed text encoder), we get the pseudo paired text embedding $\widehat{\vv}_j$  of the image $\xv_j$ as
$ \widehat{\vv}_j = \widehat{\mathbf{B}} \widehat{{\vv}}_j' = \widehat{\mathbf{B}} {\vv}_j' \approx \widehat{\mathbf{B}} \mathbf{B}^{\dagger} \uv_j$.  We note that $\widehat{\mathbf{B}} \mathbf{B}^{\dagger} \uv_j \equiv \uv_j$ when we do not reduce the projected dimension ($\widehat{d} = d$) and keep $\widehat{\mathbf{B}} = \mathbf{B}$. By replacing the text embeddings $\vv_j$ and $\widehat{\vv}_j$ in Eq.~\ref{eqn:mscore} with pseudo text embeddings $\uv_j$ and $\widehat{\mathbf{B}} \mathbf{B}^{\dagger} \uv_j$ respectively, we get the pseudo VL score matrices as:
\begin{gather}
    \mathbf{S}^{p\text{-}vl}_{i,j} = s(\uv_i, \uv_j), \quad \widehat{\mathbf{S}}^{p\text{-}vl}_{i,j} = s(\widehat{\uv}_i, \widehat{\mathbf{B}} \mathbf{B}^{\dagger} \uv_j) ~\label{eqn:pmscore}
\end{gather}
with which we define pseudo-VL Score Distillation Loss as:
\begin{align}
        \Lcal_{p\text{-}vl} =  \Lcal_{KL}(\widehat{\mathbf{S}}^{p\text{-}vl}; \mathbf{S}^{p\text{-}vl}, \mu^{p\text{-}vl}). \label{eqn:pmscore_loss}    
\end{align}

Some uni-modal self-supervised learning (SSL) works~\cite{sohn2016improved, oord2018representation} also compute the similarity score matrix (similar to $\mathbf{S}^{p\text{-}vl}$) from the same image batch, and then assign the positive/negative ground-truth label for each element in the score matrix. However, in the VL distillation, we treat $\mathbf{S}^{p\text{-}vl}$ as the supplement to $\mathbf{S}^{vl}$ which further augments text embeddings.  Moreover, we use $\mathbf{S}^{p\text{-}vl}$  as the pseudo label from the teacher and minimize the discrepancy between $\widehat{\mathbf{S}}^{p\text{-}vl}$ and $\mathbf{S}^{p\text{-}vl}$ without any ground-truth labels.

\vspace{1mm}
\noindent \textbf{Uni-Modal Distance Preserving Regularizer $\Lcal_{udist}$.} In addition to matching the similarity score of a student image embedding and a teacher image embedding in $\Lcal_{p\text{-}vl}$, we introduce a regularizer $\Lcal_{udist}$, which distills similarity score $s$ of two normalized student image embeddings\footnote{The similarity score of two normalized embeddings already encodes their relative locations.} from the teacher model, to keep the geometry of image embeddings in the student model close to that in the teacher model.  

Suppose we have two images $\xv_i$ and $\xv_j$ as well as their projected embeddings ($\uv_i$ and $\uv_j$) in the teacher's feature space and projected embeddings  ($\widehat{\uv}_i$ and $\widehat{\uv}_j$) in the student's feature space . We define the score matrices to preserve the distances of image embeddings as:
\begin{equation}~\label{eqn:udist_s}
    \mathbf{S}^{udist}_{i,j} = s(\uv_i, \uv_j), \quad \widehat{\mathbf{S}}^{udist}_{i,j} = s(\widehat{\uv}_i, \widehat{\uv}_j).
    \vspace{-2mm}
\end{equation}
We define the uni-distance preserving loss as a regularization term in the VL distillation as:
\begin{align}
      \Lcal_{udist} =  \Lcal_{KL}(\widehat{\mathbf{S}}^{udist}; \mathbf{S}^{udist}, \mu^{udist}). \label{eqn:udist_loss}
\end{align}
Although $\mathcal{L}_{udist}$ and $\mathcal{L}_{p\text{-}vl}$ only differ in $\widehat{\mathbf{B}} \mathbf{B}^{\dagger} u_j$ and $\widehat{u_j}$, their mechanisms are  theoretically different: $\mathcal{L}_{p\text{-}vl}$ views image features as pseudo text features and utilizes them for distillation (thus $\mathcal{L}_{p\text{-}vl}$ is a  text-encoder aware loss), while $\mathcal{L}_{udist}$ simply preserves the geometric structure in the visual encoder (thus $\mathcal{L}_{udist}$ is a text-encoder agnostic regularizer).

\subsection{Constructing Visually-Grounded Text Corpus}~\label{sec:choose_text}
To effectively distill the information from the pre-trained VLFMs, the choice of image corpus $\Xcal$ and text corpus $\Tcal$ is crucial. Constructing image corpus $\Xcal$ is relatively easy due to large scale natural images available on web although care must be taken to filter them to avoid duplicates and harmful content and increase diversity. However, we cannot simply use text crawled from web as $\Tcal$, because the concept distribution of natural language corpus is very different from that of a visual-grounded sentence corpus. As we show in Sec.~\ref{sec:analyze_text_choose}, we use 3 million unfiltered natural sentences as $\Tcal$ which gives much worse performance than using 3 million image captions of GCC-3M~\cite{sharma2018conceptual}. 
So it is important to select $\Tcal$ relating to visual concepts. 

With an image-text paired dataset $\{(\xv_i, \tv_i)\}_{i=1}^N$,  a simple option is that we take $\Xcal=\{\xv_i\}_{i=1}^N$ and $\Tcal=\{\tv_i\}_{i=1}^N$, where $\Tcal$ and $\Xcal$ have overlapped semantic meanings.
However, we do not assume the availability of any image-text paired data and this simple option is not achievable . 

Since the vision-language teacher model $[f_\theta, g_\phi, \mathbf{A}, \mathbf{B}]$ maps images and text into the same feature space, we can quantify the modality gap between $\Tcal$ and $\Xcal$ by measuring the distribution discrepancy between their projected embedding distributions. Given a large NLP corpus $\Tcal_{large}$, we can select $\Tcal$ from $\Tcal_{large}$, by minimizing the discrepancy between $\Tcal$'s and $\Xcal$'s embedding distributions:
\begin{align}
    \min_{\Tcal \subset \Tcal_{large}} \quad & \text{Discrepancy}(\Ucal, \Vcal) \label{eqn:Tcaloptim} \\
    \text{s.t.}\quad &\Ucal = \{\mathbf{A} f_{\thetav}(\xv):\xv \in \Xcal\}, \label{eqn:Uspace} \\
    & \Vcal = \{\mathbf{B}  g_{\phiv}(\tv):\tv \in \Tcal\} \label{eqn:Vspace}
\end{align}
This is a combinatorial optimization problem. It is expected to be NP-hard to find the exact global minimum. We propose  Algorithm~\ref{alg:Tcal}  using greedy search to approximately solve the problem.  Specifically, For each image in image datasets, we select the sentence with the highest similarity score (computed by the teacher) from NLP Corpus. We then form a visual-grounded text corpus with the selected text.  We assume that the cardinality of $\Tcal$ and $\Ucal$ is similar. If we want $|\Tcal| < |\Ucal|$, we can simply do a $|\Tcal|$-mean clustering of Algorithm~\ref{alg:Tcal}'s outputs, and construct $\Tcal$ with the resulting cluster centers. We do not see a need for $|\Tcal| > |\Ucal|$, since the image corpus $\Xcal$ can be as large as we want.

Generally, the downstream tasks are unknown before distillation. We use our constructed $\Tcal$ as the text input and call this \textbf{Task-Agnostic VL Distillation}. However, in practice, sometimes we know some of the class names used in the downstream tasks before distillation. In this case, we can incorporate those class names into the training text corpus $\Tcal$. We refer to this as \textbf{Task-Aware VL Distillation}. We compare these two VL distillations in Sec~\ref{sec:task_aware_task_agnostic}.

\begin{algorithm}
\small 
  \KwIn{image embeddings $\Ucal$ as defined in Eq.\ref{eqn:Uspace}. A large text corpus $\Tcal_{large}$.}
  \KwOut{Selected text corpus $\Tcal$, and $|\Tcal| \approx |\Ucal|$}
  $\Ucal_{left} \longleftarrow \Ucal$, \, $\Tcal_{avail} \longleftarrow \Tcal_{large}$, \, $\Tcal \longleftarrow \emptyset$, $U_p = \infty$\;
  \While{$\Ucal_{left} \neq \emptyset$ and $|\Ucal_{left}|/U_p <0.95$ }{
    $U_p = |\Ucal_{left}|$, \ $Matched = dict()$\;
    \For{$\uv \in \Ucal_{left}$}{
      \tcc{find the best text that matches the image}
      $\tv(\uv) = \argmax\limits_{\tv \in \Tcal_{avail}} s(\uv,\mathbf{B}\cdot g_{\phiv}(\tv)) $ \tcp*\;
      $Matched[\uv] = \tv(\uv)$\;
    }
    \For{$\uv, \tv \in Matched.items()$}{
      \tcc{For all images matching to the same text, pick the first match}
      \If {$\tv \in \Tcal_{avail}$}{
          $\Ucal_{left} \longleftarrow \Ucal \backslash \{\uv\}$\; 
          $\Tcal_{avail} \longleftarrow \Tcal_{avail} \backslash \{\tv\}$, \, $\Tcal.add(\tv)$\;
        }
    }
  }
  \caption{Constructing text corpus $\Tcal$}
  \label{alg:Tcal}
\end{algorithm}

\normalsize 

\vspace{-10pt}

%% file: 4_experiments.tex
\input{tables/sota.tex}

\section{Experiments}~\label{sec:experiment}
We first compare our distilled models to two state-of-the-art VLFMs with the same model capacity, CLIP~\cite{radford2021learning} and UniCL~\cite{yang2022unified}. We then compare task-agnostic and task-aware knowledge distillation, and investigate the influence of data scale on transferability and robustness. Finally, we carefully ablate our proposed distillation losses and our algorithm for text corpus construction. 

\subsection{Settings} ~\label{sec:setting} 

\vspace{-6mm}
\noindent \textbf{Evaluation benchmarks.} Foundation models are typically evaluated on transferability to downstream tasks (via zero-shot and linear probing) as well as robustness to data shifts. 
Following \cite{li2022elevater}, we evaluate all baselines and our models in three settings: (1)~Average \textbf{Zero-Shot on ELEVATER}~\cite{li2022elevater}, a dataset of 20 image-classification tasks; (2)~\textbf{Zero-Shot on IN-1K}, the ImageNet-1K~\cite{imagenet_cvpr09} validation set; (3)~Average \textbf{Linear Probing on ELEVATER}.  For robustness, we follow CLIP~\cite{radford2021learning} to report  average zero-shot performance on five datasets~\cite{recht2019imagenet,hendrycks2021many,barbu2019objectnet, wang2019learning, hendrycks2021natural} with domain shifts from IN-1K.

\vspace{1mm}
\noindent \textbf{Training Data.} Following the academic track proposed in ELEVATER~\cite{li2022elevater}, we form our \textit{image corpus} with images from ImageNet-21K (i.e. ImageNet-22K~\cite{kolesnikov2020big} excluding IN-1K classes), GCC-15M (including GCC-3M~\cite{sharma2018conceptual} and GCC-12M~\cite{changpinyo2021conceptual}) and YFCC-14M~\cite{thomee2016yfcc100m}. \textbf{We construct our \textit{text corpus} in two different ways}: 
(1) Following UniCL~\cite{yang2022unified}, from GCC-15M and YFCC-14M captions and the prompt sentences with ImageNet-21K (IN-21K) class names and 80 templates;
(2) Selecting $\Tcal$ from $\Tcal_{large}$ using images in GCC-15M and YFCC-14M with Algorithm~\ref{alg:Tcal}. We use ROBERTa~\cite{liu2019roberta}'s pretraining datasets \cite{Hamborg2017, zhu2015aligning, Gokaslan2019OpenWeb, trinh2018simple, wikidump} (total of 1.58B sentences) as  $\Tcal_{large}$.

Note that we generally do not use paired image-text data in training. We never load image-text pairs and never use pair labels explicitly in our loss function unless specified. For each experiment, we specify the exact image and text corpora used for distillation. 

\noindent \textbf{Other Settings.} We find that $\Lcal_{vl}$ alone achieves good performance, so we use it as our loss function in most experiments except for Table~\ref{table:compare_with_clip} \& \ref{table:task_agnostic_vs_task_aware} and Fig.~\ref{fig:loss_ablation}. In all experiments we distill only the image encoder and use it together with the teacher's text encoder in evaluation. See Supplementary Material Sec.A for implementation and evaluation details.

\subsection{Comparison with CLIP and UniCL}~\label{sec:compare_with_sota}

\vspace{-6mm}
\noindent \textbf{Comparison with CLIP.} We distill a small model from the released CLIP-ViT-L/14 checkpoint using the ViT-B/32 image encoder~\cite{dosovitskiy2020image}. We compare \smallclip with CLIP-ViT-B/32 in Table~\ref{table:compare_with_clip}. Both models have the same inference cost (4.4 G FLOPs/img), but  CLIP is trained on the private 400M WiT dataset~\cite{radford2021learning}, while ours uses 40M images and 28.6M sentences from public datasets (IN-21K, GCC-15M and YFCC-14M). Training with just $\Lcal_{vl}$ slightly underperforms CLIP, but after adding $\Lcal_{p\text{-}vl}$ to expand the vocabulary, the two models' performance becomes similar across the zero-shot and linear-probing testbeds.  $\Lcal_{udist}$ improves our zero-shot accuracy on IN-1K and linear-probing on ELEVATER but reduces zero-shot accuracy on ELEVATER. 
The robustness score of our distilled model is higher than CLIP-ViT-B/32 when training with $\Lcal_{p\text{-}vl}$ and $\Lcal_{udist}$. 
While this can be partially explained by our higher accuracy on IN-1K, it is still remarkable as we use less than one-tenth of CLIP training data and no image-text pairs.  

Instead of captions, we also try using a text corpus $\Tcal$ consisting of 28.4M sentences selected using Algorithm~\ref{alg:Tcal} from a language-only corpus $\Tcal_{large}$, using query images from GCC-15M and YFCC-14M. Distilling on $\Tcal$ and IN-21K prompt sentences, Distill-ViT-B/32 yields better average zero-shot performance on ELEVATER and IN-1K than CLIP-ViT-B/32 (61.4\% vs. 60.3\%), better linear probing performance (79.2\% vs. 78.2\%)  as well as better robustness ($50.2\%$ vs. $48.6\%$).   We analyze the quality of our constructed $\Tcal$ and the human-annotated captions in Sec.~\ref{sec:analyze_text_choose}. 

We note that \smallclip falls short on Zero-Shot on ELEVATOR. After the careful analysis, we find the large CLIP-ViT-L/14 performs much worse than the small CLIP-ViT-B/32 on PatchCamelyon~\cite{veeling2018rotation} (51.2\% vs. 60.7\%) and KITTI Distance~\cite{fritsch2013new} (13.8\% vs. 29.0\%) in ELEVATER. After removing these two tasks, \smallclip yields the same zero-shot score (61.0\%) on ELEVATER as CLIP-ViT-B/32. See Supplementary Material Sec. E for more analysis for each individual downstream dataset. 

\begin{figure*}
\begin{center}
     \includegraphics[width=\linewidth]{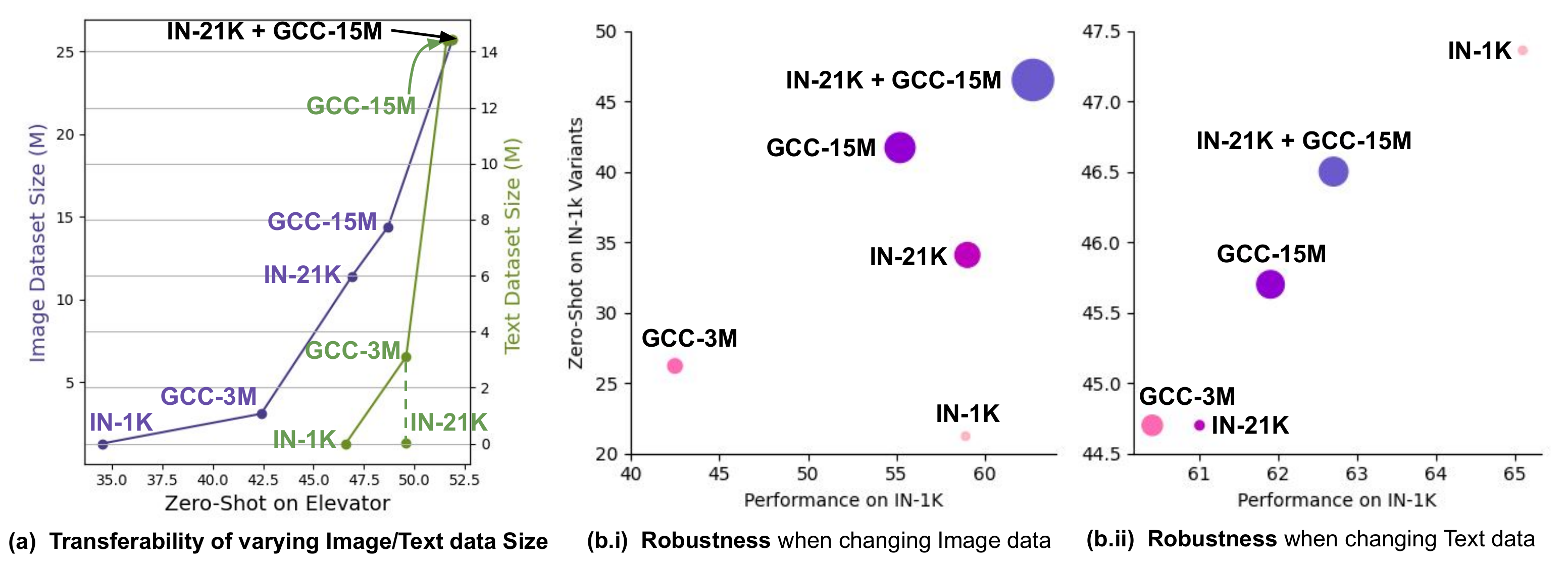}
\end{center} 
\vspace{-15pt}
   \caption{\small 
   \textbf{Transferability and Robustness for different Image/Text Dataset Sizes.}  (a) zero-shot transferability of our student model increases with larger training image/text corpus; (b.i) shows robustness strongly correlates to the training image dataset size (represented as the dot size); (b.ii) shows robust score strongly correlates to IN-1K performance when changing the training text.} 
   \label{fig:change_dataset_size} 
\end{figure*}

\vspace{1mm}
\noindent \textbf{Comparison with UniCL.} In Table~\ref{table:compare_with_unicl}, we compare our distillation approach with UniCL~\cite{yang2022unified}, which trains contrastively on smaller-scale public image-text pairs, unifying captioning datasets and pseudo-captioned classification datasets. Following the settings in UniCL, we adopt \enquote{IN-21K} and \enquote{IN-21K + YFCC14M} as our training datasets and use Swin-Tiny Transformer~\cite{liu2021Swin} as our student image encoder. In UniCL, both image and text encoders are trained from scratch. We report UniCL's performance by evaluating its released checkpoints trained with two different data sources. For a fair comparison, we further introduce UniCL* in which we use the pretrained CLIP-ViT-L/14 text encoder as UniCL's. During training, we fix the text encoder's weights and only optimize the image encoder with contrastive loss.  UniCL* achieves better zero-shot performance than UniCL due to CLIP's strong text encoder. Nevertheless, our Distilled-UniCL* significantly outperforms UniCL* on all evaluation benchmarks with only the $\Lcal_{vl}$ loss. This indicates that distilling a small VLFM using strong pseudo-labels from large VLFMs is better than contrastive pretraining when we do not have large-scale datasets. Even though our experiment with \enquote{IN-21K + YFCC-14M} shows that enlarging data scale reduces the performance gap between  distillation and pretraining (more analysis in Supplementary Material Sec. E), \ours is more data-efficient, since it does not require any expensive image-text pairs.

\subsection{Task-Agnostic vs. Task-Aware}~\label{sec:task_aware_task_agnostic}

\input{tables/task_aware.tex} 

\vspace{-5mm}
We evaluate Task-agnostic VL Distillation and Task-aware VL Distillation in Table~\ref{table:task_agnostic_vs_task_aware}. We show performance on downstream tasks with known classes and generalization to other downstream tasks with unknown classes, under different loss weight $\lambda_1$.
BeamCLIP~\cite{kimtransferring} uses \textit{only} the IN-1K prompt text to distill CLIP's image encoder. Table~\ref{table:task_agnostic_vs_task_aware} shows that this generalizes poorly to other unknown downstream tasks (\eg ELEVATER). 
With larger weight $\lambda_1$ on $\Lcal_{p\text{-}vl}$ to expand text embeddings, BeamCLIP's student model generalizes better on ELEVATER but is still worse than our task-agnostic knowledge distillation.
When we target multiple downstream tasks and only use prompt text with their class names (\ie IN-1K and ELEVATER) as the input text corpus (denoted as \enquote{DS Prompt Text} in Table~\ref{table:task_agnostic_vs_task_aware}), it is hard to balance different tasks, \eg zero-shot performance on ELEVATER improves while zero-shot performance on IN-1K worsens compared to \cite{kimtransferring}. Combining the large text corpus $\Tcal$ and the prompt sentences of downstream class names is a good practice for task-aware distillation.

\subsection{Influence of Dataset Scale}~\label{sec:dataset_scale}
We investigate the influence of image and text datasets' scale on transferability and robustness of the student model by fixing the dataset scale of one modality and varying the other. For the fixed-size modality, we use images or text from \enquote{IN-21K + GCC-15M}.  

From Fig.~\ref{fig:change_dataset_size}~(a), we find that the transferability of student foundation models improves with larger image or text corpus, but it is more sensitive to the image corpus size. Also, prompt sentences with IN-21K class names describe diverse visual concepts, so training with these achieves comparable transferability to training with 3M captions from GCC-3M. 

In Fig.~\ref{fig:change_dataset_size}~(b), we study the correlation between robustness on IN-1K variant datasets and original performance on IN-1K, as well as the correlation between robustness and the size of image/text corpus.  Fig.~\ref{fig:change_dataset_size}~(b.i) shows that when we fix the text corpus size, robustness correlates with the training image corpus size more strongly than with  IN-1K performance. Even though distilling directly with IN-1K images produces better performance on IN-1K, it does not guarantee better robustness to domain shifts from IN-1K. In Fig.~\ref{fig:change_dataset_size}~(b.ii), we freeze the image corpus size and find a different trend, in which robustness directly relates to performance on IN-1K regardless of the text corpus size. 

We conclude that VL distillation methods should focus on increasing the training image set to achieve better transferability and robustness. If downstream class names are unknown, it is critical to construct a text corpus that covers more visual concepts. If downstream class names are known, using them during distillation greatly benefits robustness.

\begin{figure}
\begin{center}
     \includegraphics[width=0.95\linewidth]{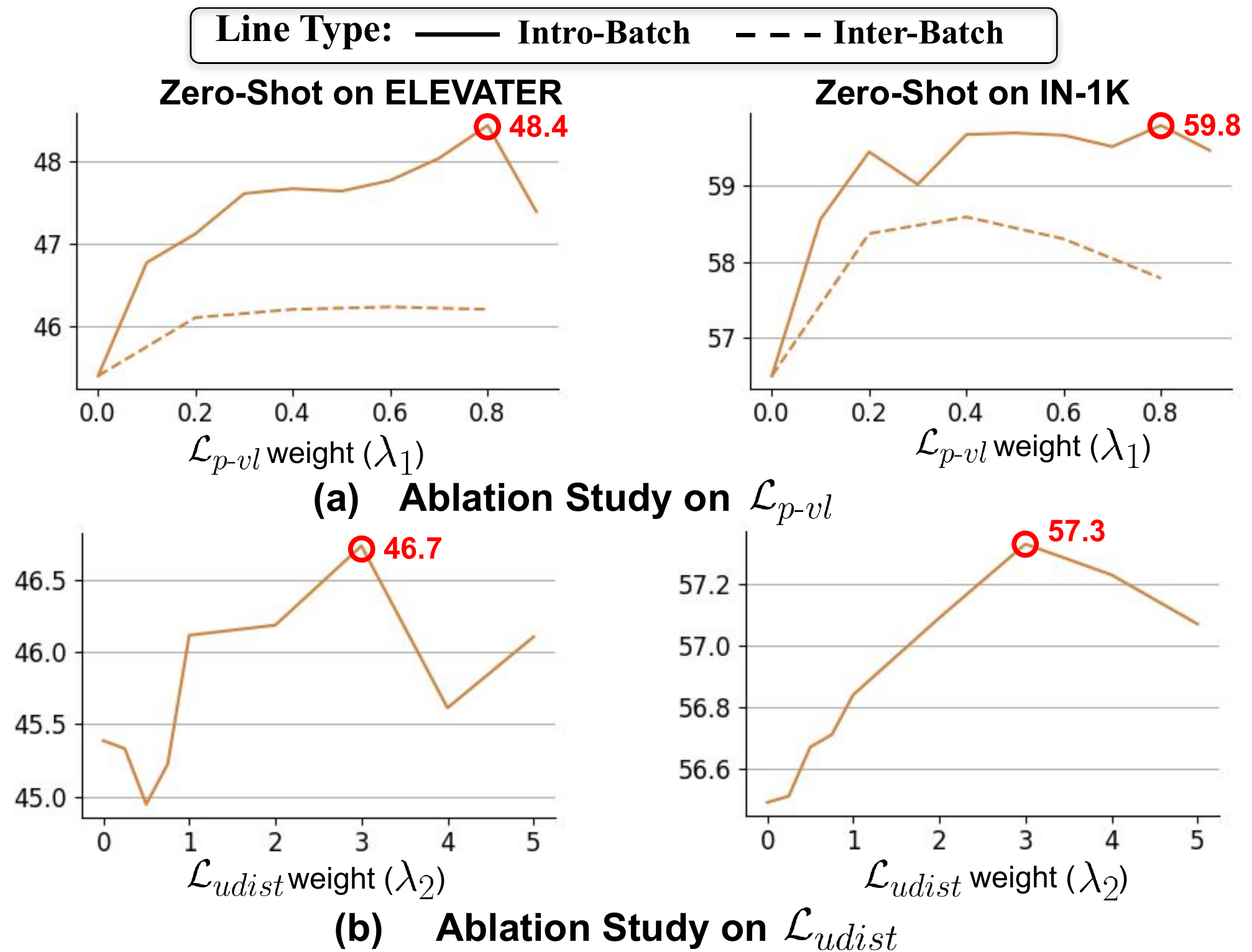}
\end{center} 
\vspace{-10pt}
    \caption{\small 
   \textbf{Ablation Studies on $\mathcal{L}_{p\text{-}vl}$ and $\mathcal{L}_{udist}$.} 
   }    
   \label{fig:loss_ablation} 
   \vspace{-20pt}
\end{figure}

\subsection{Ablation Studies on Losses}~\label{sec:ablation_on_losses}
We examine the effect of $\Lcal_{p\text{-}vl}$ and $\Lcal_{udist}$ by Zero-Shot on ELEVATER and IN-1K with  IN-21K images and part of GCC-3M captions as the training data.  See Supplementary Material Sec.~G for more ablation studies on losses.

\noindent \textbf{Ablation on $\mathcal{L}_{p\text{-}vl}$.} We gradually put more weights on the $\mathcal{L}_{p\text{-}vl}$  by increasing $\lambda_1$ in Eq.~\ref{eq:total_loss} from 0 to 1 in Fig.~\ref{fig:loss_ablation}~(a).  We compare the inter-batch version (i.e. $\uv_i$ and $\uv_j$ in Eq.~\ref{eqn:pmscore}  from different batches) and intro-batch version (i.e. $\uv_i$ and $\uv_j$ from the same batch) of $\Lcal_{p\text{-}vl}$ and find the intro-batch $\Lcal_{p\text{-}vl}$ performs better than inter-batch $\Lcal_{p\text{-}vl}$ , so we keep intro-batch version in other experiments. Furthermore, adding $\Lcal_{p\text{-}vl}$ with $\lambda_1 \le 0.9$ brings  better zero-shot performance than only using $\Lcal_{vl}$ in all three settings. However, we observe the dramatic performance drop when we totally replace $\Lcal_{vl}$ with $\Lcal_{p\text{-}vl}$ (\ie $\lambda_1=1$). We argue improvement with smaller $\lambda_1$'s and drop at $\lambda_1 =1$ both due to the gap between images and text embeddings in the shared feature space (More analysis in Sec.~\ref{sec:distribtion_analysis}.).

\noindent \textbf{Ablation on $\mathcal{L}_{udist}$.} We increase $\lambda_2$ from 0 to 5, to introduce $\Lcal_{udist}$ as a regularization term. Generally, $\Lcal_{udist}$ benefits the Zero-Shot on ELEVATER since it tries to preserve the geometry of image features. 
$\Lcal_{udist}$ slightly improves IN-1K performance when $\lambda_2$ is small but it quickly harms IN-1K performance when $\lambda_2$ gets larger. We suspect the poor student embedding in early training along with the large regularization term detours the gradient decent trajectory. 
We find $\Lcal_{udist}$ is less effective than the similar $\Lcal_{p\text{-}vl}$, so we only use  $\Lcal_{udist}$ as a regularizer.
Our main experiment (Table.~\ref{table:compare_with_clip}) further shows $\Lcal_{udist}$ is less effective when applying $\Lcal_{p\text{-}vl}$ and $\Lcal_{udist}$ together.

\subsection{Analysis of Constructed Text Corpus}~\label{sec:analyze_text_choose} 

\vspace{-6mm}
\noindent \textbf{Text Corpus based on GCC-3M~\cite{sharma2018conceptual} Images.} We first compare the distillation performance of our constructed $\Tcal$ with the original GCC-3M captions and with randomly sampled NLP sentences in Table~\ref{table:cc3m_different_text_corpora}. $\Tcal$ is constructed using Algorithm~\ref{alg:Tcal} on the large-scale NLP corpus and the GCC-3M image set. We find that our constructed $\Tcal$ yields better zero-shot and compatible linear-probing performance compared to the original GCC-3M Captions, while the unfiltered NLP corpus at the same size performs poorly.
\input{tables/cc3m_select_sentence.tex}

\begin{figure}[t]
\begin{center}
     \includegraphics[width=0.9\linewidth]{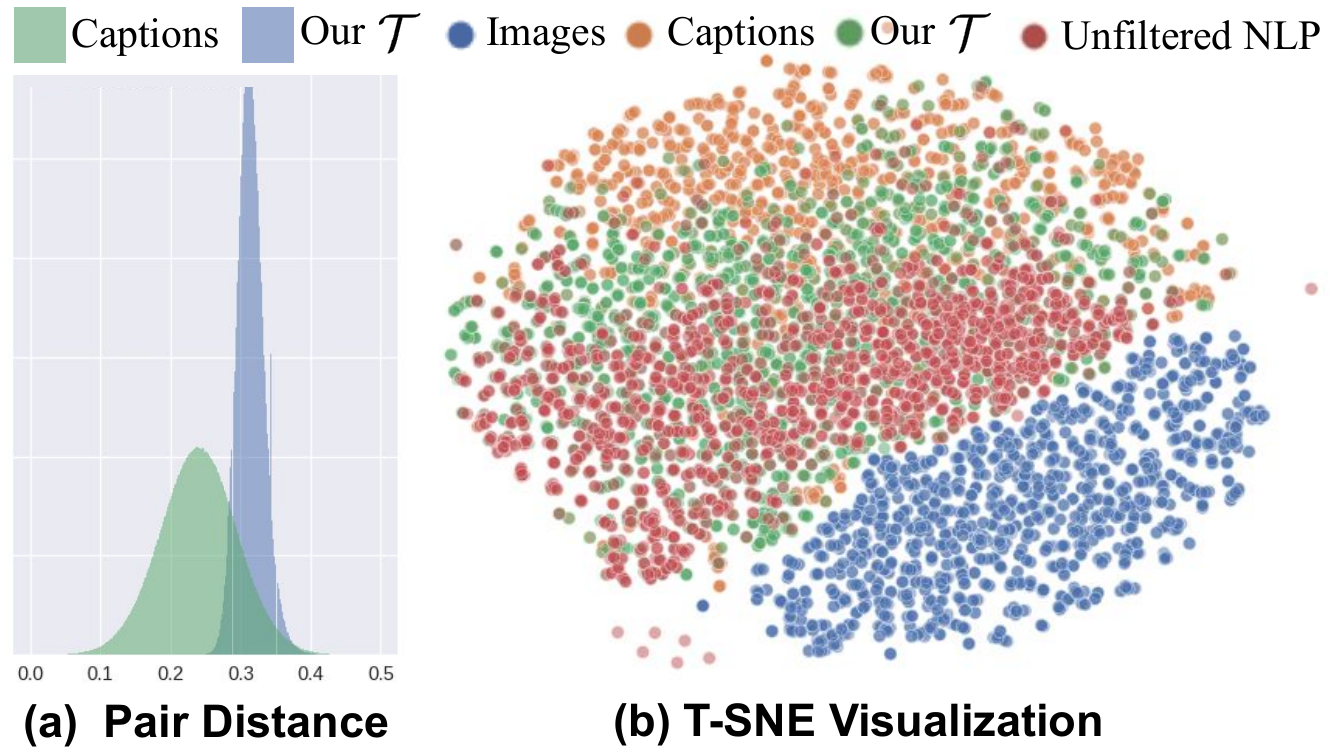}
\end{center} \vspace{-15pt}
   \caption{\small 
   \textbf{ Analysis of Text Corpus selected from the ROBERTa NLP Corpus.}
Best viewed in color. 
   }
   \label{fig:text_corpus}
   \vspace{-15pt}
\end{figure}

\noindent \textbf{Pair-level Analysis.} 
We analyze the quality of image-text pairs via the similarity score computed with CLIP-ViT-L/14 (teacher model). In Fig.~\ref{fig:text_corpus}~(a), we compute the histogram of similarity scores for image-caption pairs in GCC-15M and YFCC-14M. We also compute the same similarity score histogram for our selected sentence with its query image. Our  sentences selected from 1.5B candidate sentences yield  higher average similarity score than the human annotations. 
 See our visualization for the selected text and the effect of each NLP dataset in Supplementary Material Sec. B-C. 

\noindent \textbf{Distribution-level Analysis.}~\label{sec:distribtion_analysis} We also analyze the distribution of our constructed $\Tcal$ in the shared feature space. We plot T-SNE (see Fig.~\ref{fig:text_corpus}~(b)) of normalized embeddings for samples from four different corpora: images,  human-annotated captions,  our selected sentences and random sentences from the NLP corpus. Even though original CLIP models~\cite{radford2021learning} yield astonishing zero-shot performance on a large variety of downstream tasks, the images and their human annotated captions surprisingly do not overlap in the T-SNE visualization (We also provide MMD among these four corpora in Supplementary Material Sec. F). We conclude that the contrastive loss used in CLIP only pushes the text closer to its related image does not close the distribution gap between image and text corpora in feature space. This explains the effectiveness of $\Lcal_{p\text{-}vl}$ where we use the image embeddings ({\color{blue1} Blue} dots in Fig.~\ref{fig:text_corpus}~(b)) as the pseudo text embeddings. $\Lcal_{p\text{-}vl}$ expands the text feature space and unsurprisingly leads to better performance. When we completely replace $\Lcal_{vl}$ with $\Lcal_{p\text{-}vl}$, the distillation performance drops a lot due to the large gap between the image and text modalities in the feature space. Moreover, the distributions of our selected $\Tcal$ and the human-annotated captions are more similar. On the other hand, samples of the ROBERTa NLP Corpus have a different distribution from the visually-grounded sentences. These results provide further support for our Algorithm~\ref{alg:Tcal}.

%% file: tables/sota.tex
\begin{table*}[h]
    \vspace{-5pt}
    \begin{center}
        \resizebox{\linewidth}{!}{
        \begin{tabular}{c c c  c c c c }
            \Xhline{3\arrayrulewidth} 
             \multirow{2}{*}{Method} & \multirow{2}{*}{Train Data} &  \multirow{2}{*}{Loss} &  \multicolumn{2}{c}{Zero-Shot}  & Linear Probing & \multirow{2}{*}{Robustness} \\
            & & & ELEVATER  & IN-1K &  ELEVATER \\
            \Xhline{3\arrayrulewidth}
            CLIP-ViT-B/32  & 400M Image-Text Pairs & Contrastive Loss & \textbf{57.2\%} & 63.4\% & 78.2\% & 48.6\% \\
            \hline
            \multirow{4}{*}{\makecell{Distill-ViT-B/32 \\ (\ours) }}   & \multirow{3}{*}{\makecell{ 40M Images, 28.6M Captions}}  & \multirow{1}{*}{$\mathcal{L}_{vl}$} &  53.5\%  & 64.2\% & 77.9\%  & 48.1\% \\
             & & \multirow{1}{*}{$0.7\mathcal{L}_{vl}$ + $ 0.3 \Lcal_{p\text{-}vl}$ } &  55.8\%  & 64.8\% & 78.4\% & \underline{49.4\%} \\
            & & \multirow{1}{*}{$0.7\mathcal{L}_{vl}$ + $0.3 \Lcal_{p\text{-}vl}$ +  $0.5\mathcal{L}_{udist}$ } & 55.0\%  & \underline{65.1\%} & \underline{78.6\%} & \underline{49.4\%} \\
             \cline{2-7}
            & \multirow{1}{*}{\makecell{40M Images, 28.4M NLP Text}}  &  \multirow{1}{*}{$0.7\mathcal{L}_{vl}$ + $ 0.3 \Lcal_{p\text{-}vl}$} &   \underline{56.4\%} 
            &\textbf{ 66.5\%  }& \textbf{79.2\%} &\textbf{ 50.2\% }\\
             \Xhline{3\arrayrulewidth} 
        \end{tabular}
        } 
        \vspace{-5pt}
        \caption{\small \textbf{\ours vs. CLIP.} We distill  Distill-ViT-B/32 from CLIP-ViT-L/14 (81.1G FLOPs/img) and compare it with CLIP-ViT-B/32. } \label{table:compare_with_clip}
    \end{center}
\vspace{-15pt}
\end{table*}

\begin{table}[h]
    \begin{center}
        \resizebox{\linewidth}{!}{
        \begin{tabular}{c c c c c }
            \Xhline{3\arrayrulewidth} 
           \multirow{2}{*}{Dataset} &   \multirow{2}{*}{Method} &  \multicolumn{2}{c}{Zero-Shot}  & Linear Probing \\
           &  & ELEVATER  & IN-1K &  ELEVATER\\
            \Xhline{3\arrayrulewidth} 
             \multirow{3}{*}{IN-21K}  &  UniCL & 27.2\% & 28.5\% & 74.8\%\\
             & UniCL* & 40.9\%  &  51.4\% & 75.3\%  \\
             & Distill-UniCL* &  \textbf{45.6\% }& \textbf{59.5\%} &\textbf{ 76.2\% }\\
             \Xhline{3\arrayrulewidth} 
             \multirow{3}{*}{\makecell{IN-21K + \\ YFCC-14M}}  & UniCL & 37.1\% & 40.5\% & \textbf{77.1\%}\\
             & UniCL* & 44.6\% & 58.7\%  & 75.4\% \\
             & Distill-UniCL* &  \textbf{47.6\%} & \textbf{ 60.0\% }& 76.6\%  \\
             \Xhline{3\arrayrulewidth} 
        \end{tabular}
        } 
        \vspace{-5pt}
        \caption{\small \textbf{\ours vs. UniCL.}  We distill a Swin-Tiny Transformer from CLIP ViT-L/14 and compare it to Swin-Tiny UniCL model trained with the same dataset. UniCL* is defined in Sec.~\ref{sec:compare_with_sota}. }\label{table:compare_with_unicl}
    \end{center}
\vspace{-15pt}
\end{table}

%% file: tables/task_aware.tex
\begin{table}
    \begin{center}
        \resizebox{\linewidth}{!}{
        \begin{tabular}{c| c c  c c }
            \Xhline{3\arrayrulewidth} 
            \multirow{2}{*}{$\lambda_1$} &  \multirow{2}{*}{Input Text Corpus} &  \multicolumn{2}{c}{Zero-Shot}  & Linear Probing \\
            & & ELEVATER  & IN-1K &  ELEVATER \\
            \Xhline{3\arrayrulewidth}
         \multirow{4}{*}{0} & \cellcolor{lightgray!15} \textit{28.6M Text} & \cellcolor{lightgray!15} 53.5\% & \cellcolor{lightgray!15} 64.2\% & \cellcolor{lightgray!15} 77.9\%  \\ 
           & IN-1K Prompt Text\cite{kimtransferring} & 47.4\% & \textbf{67.2\%}&  76.9\% \\
           &  DS Prompt Text  & \underline{56.8\%} & 57.2\% & \underline{78.9\%}\\
           & \textit{28.6M}  +  DS Prompt Text & \textbf{ 57.5\%} & \underline{65.6\%} &\textbf{ 79.2\% }\\
       \Xhline{3\arrayrulewidth}
          \multirow{4}{*}{\makecell{0.3}} & \cellcolor{lightgray!15}\textit{ 28.6M Text} &  \cellcolor{lightgray!15} 55.8\% & \cellcolor{lightgray!15} 64.8\% & \cellcolor{lightgray!15} 78.4\% \\ 
           & IN-1K Prompt Text\cite{kimtransferring} & 50.8\% & \textbf{66.6\%} & 77.1\%\\
           & DS Prompt Text & \textbf{57.8\%} & 60.3\% &\textbf{ 79.4\%} \\
           & \textit{28.6M}  + DS Prompt Text  & \underline{57.7\%} & \underline{66.1\%} &\textbf{ 79.4\%} \\
         \Xhline{3\arrayrulewidth}
           \multirow{4}{*}{\makecell{0.8}} & \cellcolor{lightgray!15} \textit{28.6M Text} & \cellcolor{lightgray!15} 55.5\% & \cellcolor{lightgray!15} 64.2\% & \cellcolor{lightgray!15}  78.8\% \\ 
           & IN-1K Prompt Text\cite{kimtransferring} & 53.1\% & \textbf{65.4\%} & 78.0\%\\
           & DS Prompt Text &\textbf{ 59.4\%} & 61.7\% & \textbf{79.8\% }\\
           & \textit{28.6M}  + DS Prompt Text & \underline{57.5\%} & \underline{64.9\%} & \underline{79.5\%} \\
             \Xhline{3\arrayrulewidth} 
        \end{tabular}
        } 
        \caption{\small \textbf{Task-Agnostic vs. Task-Aware.} DS = Downstream.
        } \label{table:task_agnostic_vs_task_aware}
    \end{center}
\vspace{-12pt}
\end{table}

%% file: tables/cc3m_select_sentence.tex
\begin{table}
    \begin{center}
        \resizebox{0.95\linewidth}{!}{
        \begin{tabular}{c c c c }
            \Xhline{3\arrayrulewidth} 
            \multirow{2}{*}{Text Corpus} &  \multicolumn{2}{c}{Zero-Shot}  & Linear Probing \\
            & ELEVATER  & IN-1K &  ELEVATER\\
            \Xhline{3\arrayrulewidth} 
             GCC-3M (Text) & \underline{38.6\%} & \underline{39.0\%} & \textbf{68.2\% }\\ 
             Unfiltered NLP (3M)  & 35.9\% & 33.2\% & 65.2\% \\
             Our Constructed $\Tcal$ (3M) &\textbf{ 40.4\% }& \textbf{ 39.2\% }& \underline{67.7\%} \\
             \Xhline{3\arrayrulewidth} 
        \end{tabular}
        } 
        \vspace{-5pt}
        \caption{\small \textbf{Distillation with Different Text Corpora of the Same Size.} Images from GCC-3M serve as the image dataset. ~}\label{table:cc3m_different_text_corpora}
    \end{center}
\vspace{-5pt}
\end{table}

%% file: 5_conclusion.tex
\vspace{-2mm}
\section{Conclusion}
\vspace{-2mm}
In this paper, we propose a vision-language knowledge distillation mechanism \ours  that
distills knowledge in pre-trained VLFMs to small foundation models, without using any paired image-text data. We distill pre-trained CLIP-ViT-L/14 to our Distill-ViT-B/32 model, with only 40M public images and 28.4M unpaired text and our model rivals the CLIP-ViT-B/32 model that was pretrained on private large-scale WiT dataset in both
transferability to novel tasks and robustness to natural domain shifts. Particularly, we propose an efficient text selection algorithm and two novel distillation losses for vision-language knowledge distillation. This paper shows how to achieve a small custom foundation model with limited unpaired data and released huge CLIP foundation models, which is the first trial in distilling a multi-modal foundation model while preserving its foundation properties. There are many interesting directions not covered in this paper and left for exploration in the future, such as VL distillation with large-scale paired image-text data (\eg. 400M+) and distillation of foundation models of other multi-modalities  (\eg video-language).

%% file: A1_implementation_details.tex
\section{Implementation Details}~\label{sec:implementation_details}
\vspace{-8mm}
\subsection{Training}
We pre-compute the teacher features of all images and texts using CLIP-ViT-L/14. We apply the basic data augmentations (only random cropping, flipping and data whitening) to the input images when computing the student features. We empirically find out more advanced data augmentations (such as \enquote{rand-m9-n3-mstd0.5}) harm the distillation performance. 
We argue that the advanced data augmentations for the student model's input images enlarge the discrepancy of the student feature and the fixed pre-extracted teacher feature, which lead to the poor performance in vision-language knowledge distillation. We train all
models for 100 epochs.  During the training,  we use the Adam optimizer~\cite{kingma2014adam} with decoupled weight decay regularization~\cite{loshchilov2017decoupled}. We set the initial learning rate as $8 \times 10 ^{-4}$ and weight decay $0.05$. We warm up the training for 4 epochs and  then we decay the learning rate using a cosine schedule~\cite{loshchilov2016sgdr}.  We remove the weight decay of the weights that are the gains or biases. For our model \smallclip,  we optimize with the batch size 12288 for images and texts. For Distill-UniCL*, we adopt 8192 as the batch size. We grid search for the best hyperparameters $\mu^{vl}$, $\mu^{p\text{-}vl}$ and $\mu^{udist}$ when performing ablation studies on losses in Sec 4.5 (main paper). We finally set $\mu^{vl} = 100$, $\mu^{p\text{-}vl}=33.3$ and $\mu^{udist}=14.3$ for all experiments. 

\subsection{Evaluation}

\vspace{-2mm}
\noindent\textbf{Transferability to Novel Downstream Tasks.} We use ELEVATER toolkit~\cite{li2022elevater} to evaluate the model's zero-shot and linear probing performance on 20 image classification datasets including both coarse and fine-grained tasks: Hateful Memes~\cite{kiela2020hateful}, PatchCamelyon~\cite{veeling2018rotation}, Rendered-SST2~\cite{radford2021learning}, KITTI Distance~\cite{fritsch2013new}, FER 2013~\cite{fer2013}, CIFAR-10~\cite{krizhevsky2009learning}, EuroSAT~\cite{helber2019eurosat}, MNIST~\cite{deng2012mnist}, VOC 2007 Classification~\cite{everingham2009pascal}, Oxford-IIIT Pets~\cite{parkhi2012cats}, GTSRB~\cite{stallkamp2011german}, Resisc-45~\cite{cheng2017remote}, Describable Textures~\cite{cimpoi2014describing}, CIFAR-100~\cite{krizhevsky2009learning}, FGVC Aircraft (variants)~\cite{maji2013fine}, Food-101~\cite{bossard2014food}, Caltech-101~\cite{fei2004learning}, Oxford Flowers 102~\cite{nilsback2008automated}, Stanford Cars~\cite{krause20133d} and Country-211~\cite{radford2021learning}. Please refer to Sec.C in Supplementary Material of ELEVATER~\cite{li2022elevater} for detailed dataset statistics and evaluation metrics. We use the same prompt templates as ELEVATER toolkit. We also evaluate the zero-shot performance on ImageNet-1K~\cite{imagenet_cvpr09}. For linear-probing performance, we enable the grid search of learning rate and weight decay before finetuning the last classifier layer. 

\vspace{1mm}
\noindent\textbf{Robustness to Domain Shifts.} Following CLIP~\cite{radford2021learning}, we use five datasets which have the distribution shifts from ImageNet-1K~\cite{imagenet_cvpr09}: ImageNet-V2 Match frequency~\cite{recht2019imagenet}, ImageNet Sketch~\cite{wang2019learning}, ImageNet Adversarial~\cite{hendrycks2021natural}, ObjectNet~\cite{barbu2019objectnet} and ImageNet Rendition~\cite{hendrycks2021many}. We use the same 80 prompt templates of ImageNet-1K for these datasets and we report the average zero-shot top-1 performance on these datasets as the metric for Robustness.

%% file: A3_visualization_image_text.tex
\section{Visualization of the Constructed $\Tcal$}~\label{sec:visualization}
\begin{figure}
\begin{center}
     \includegraphics[width=1\linewidth]{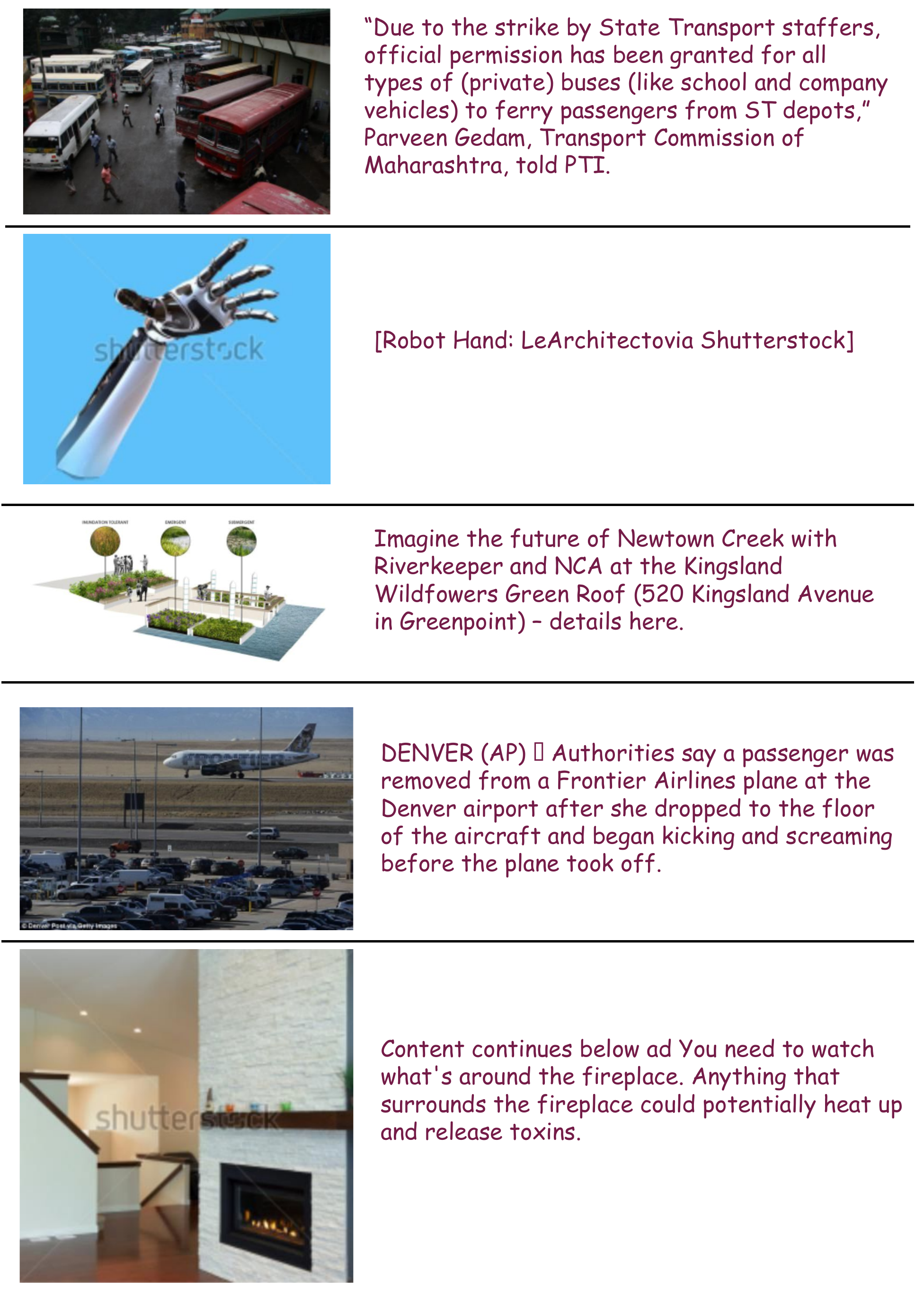}
\end{center} 
\vspace{-20pt}
   \caption{\small 
   \textbf{Selected Sentences from ROBERTa NLP Corpus. 
   }}
   \label{fig:pair_visual} 
   \vspace{-10pt}
\end{figure}

In Fig.~\ref{fig:pair_visual}, we randomly select several images and visualize their paired sentences of original human annotations and our proposed algorithms. For the second image with robot hand, the chosen sentence describes the image content more accurate than human labels. It indicates our proposed text corpus selection algorithms choose sentences which are not only close to the image in the feature space but also with reasonable concept in the visualization.

%% file: A4_contribution_NLP.tex
\section{Contribution of each NLP dataset}~\label{sec:contribution_nlp}
\begin{figure}
\begin{center}
     \includegraphics[width=1\linewidth]{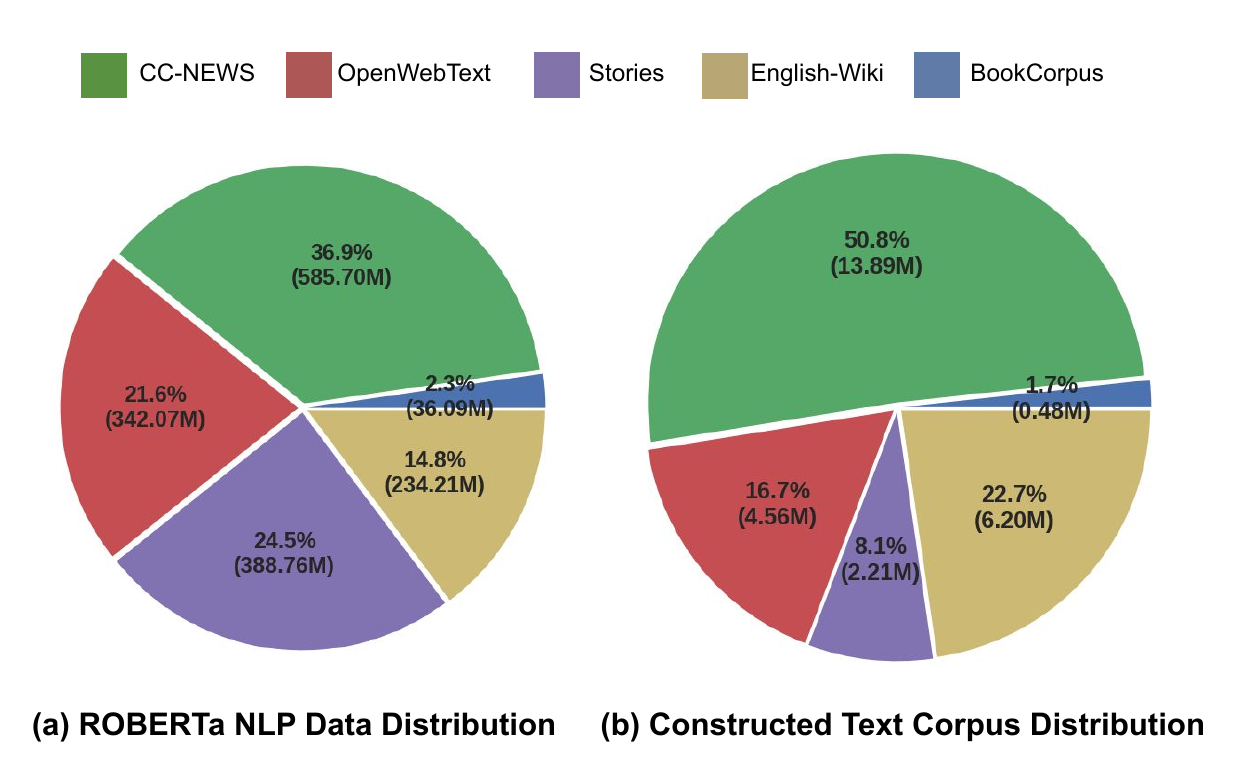}
\end{center}
\vspace{-20pt}
   \caption{\small 
   \textbf{ Analysis of Our Constructed Text Corpus from ROBERTa NLP Corpus.
Best viewed in color. 
   }}
   \label{fig:contribution_text}
   \vspace{-10pt}
\end{figure}

In Fig.~\ref{fig:contribution_text}~(a) and (b), we counted the contribution of each NLP datasets in original ROBERTa NLP Corpus~\cite{liu2019roberta} and our constructed Text Corpus with 28.4M images. Comparing with  Fig.~\ref{fig:contribution_text}~(a) and (b), we show our proposed sentence selection algorithm favors sentences from CC-NEWS~\cite{Hamborg2017} and English-Wiki, which indicates there are more visually-grounded sentences in these datasets. For instance, they might contain more visual object entities.

%% file: A5_paired_vs_unpaired_dataloading.tex
\section{Paired vs. Unpaired Dataloading}~\label{sec:dataloading}
In the main paper, we load in images and texts independently. In this section, we try to load the image and text in pairs (image-caption) with GCC-3M~\cite{sharma2018conceptual}, without introducing new losses to use the ground-truth information in the paired image and text data. We re-tune the hyper-parameter for the experiment when loading the paired data. In Table~\ref{table:pair_vs_unpair}, we show loading image and text in pairs does not benefit the transferability during the knowledge distillation and even we observe a small drop when switching to the paired dataloading mechanism. We suspect the drop results from the fewer combinations of image and text seen during the optimization if we load image and text in pairs. It would be an interesting future direction to study how to make better usage of the annotated images and its associating text  when there is a small amount of image-text data available, which is out of scope of this paper.
\input{tables/pair_vs_unpaired.tex}

%% file: tables/pair_vs_unpaired.tex
\begin{table}[h]
    \begin{center}
        \resizebox{0.8\linewidth}{!}{
        \begin{tabular}{c c c c }
            \Xhline{3\arrayrulewidth} 
            \multirow{2}{*}{Dataloading} &  \multicolumn{2}{c}{Zero-Shot}  & Linear Probing \\
            & ELEVATER  & IN-1K &  ELEVATER\\
            \Xhline{3\arrayrulewidth} 
             Unpaired & 38.6\% & 39.0\% & 68.2\% \\ 
             Paired & 38.8\% & 38.3\% & 68.0\% \\
             \Xhline{3\arrayrulewidth} 
        \end{tabular}
        } 
        \vspace{-5pt}
        \caption{\small \textbf{Paired vs. Unpaired Dataloading.}}\label{table:pair_vs_unpair}
    \end{center}
\vspace{-25pt}
\end{table}

%% file: A6_analysis_of_elevator.tex
\begin{figure*}
\begin{center}
     \includegraphics[width=1\linewidth]{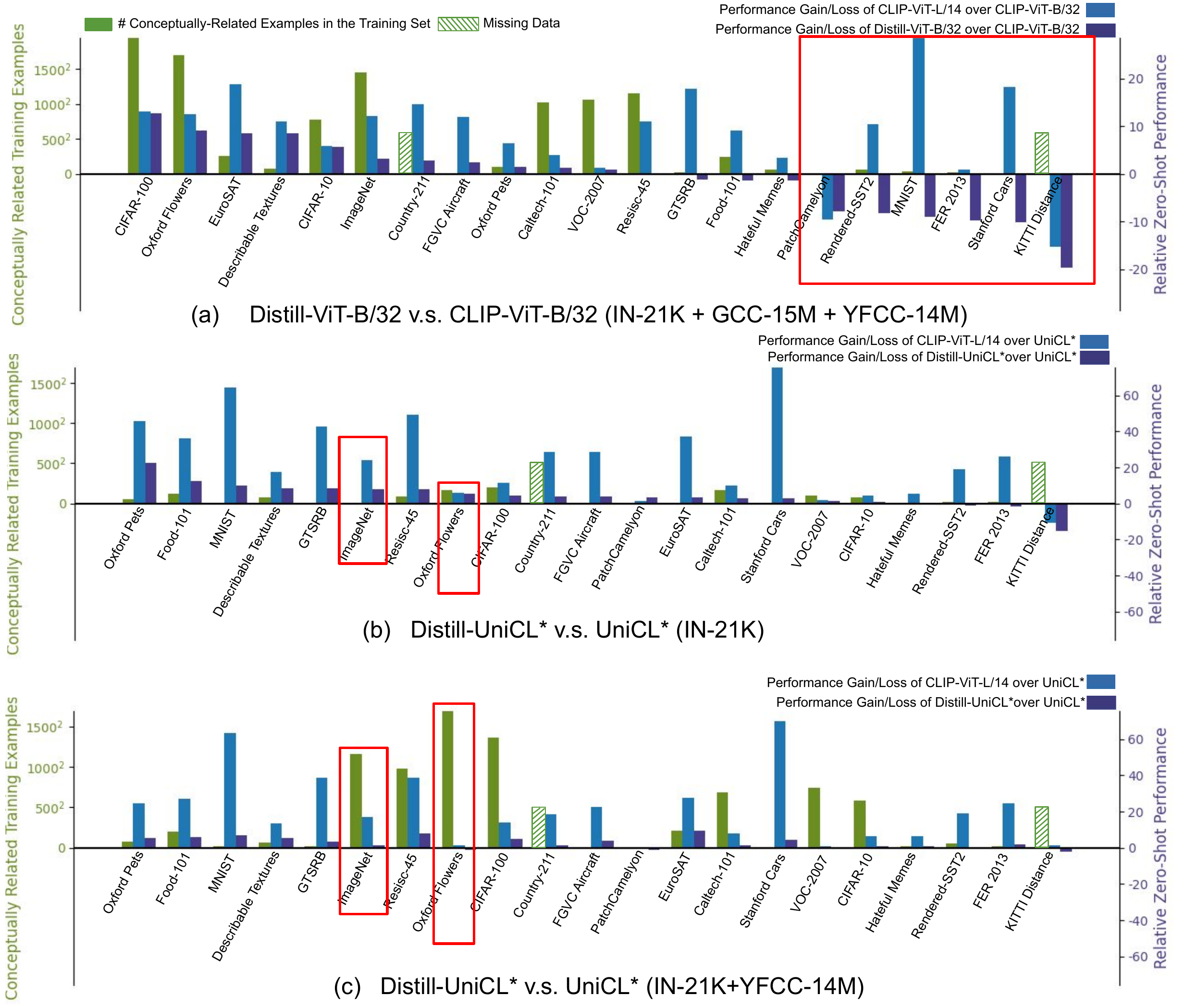}
\end{center} 
 \vspace{-20pt}
   \caption{\small 
   \textbf{ Conceptual Coverage Analysis of Training Data over Each Downstream Task.
   }}
   \label{fig:conceptual_coverage}
   \vspace{-10pt}
\end{figure*}

\section{Conceptual Coverage Analysis}~\label{sec:concept_coverage}
When we evaluate the zero-shot performance of our distilled models on various downstream tasks in ELEVATER and ImageNet-1K benchmarks, we find some tasks benefit more from the teacher model than the others. We study the performance gain or loss of our distilled model for a single downstream task with respect to the teacher model's performance and the number of conceptually-related images available during the training (see Fig.~\ref{fig:conceptual_coverage}). We compute the number of conceptually-related images to each individual downstream task  (green bars in Fig.~\ref{fig:conceptual_coverage}) from Table 11 of \cite{yang2022unified}. 
\begin{itemize}[leftmargin=*]
    \item For Fig.~\ref{fig:conceptual_coverage}~(a), we transfer the knowledge from the large teacher model CLIP-ViT-L/14 to our \smallclip and compare with CLIP-ViT-B/32. Both CLIP-ViT-L/14 and CLIP-ViT-B/32 are trained on the private 400M image-text pairs while \smallclip is trained on 40M images (consisting of images from IN-21K, GCC-15M and YFCC-14M) and 28.4M unpaired sentences. Our \smallclip has obvious worse zero-shot performance than CLIP-ViT-B/32 on five out of twenty-one datasets  (\ie PatchCamelyon, MNIST, FER-2013, Stanford Cars and KITTI Distance)  due to either lack of conceptually-related images in our small training set (MNIST, FER-2013 and Stanford Cars) or the poor performance of the teacher model (PatchCamelyon and KITTI Distance). Notably, if we do not consider PatchCamelyon and KITTI Distance datasets where the large CLIP model does not full fill a teacher role, our \smallclip achieves the average Zero-Shot performance  $61.03\%$ (vs.  $60.95\%$ for CLIP-ViT-B/32) on the remaining 18 datasets in ELEVATER benchmark.
    \item For Fig.~\ref{fig:conceptual_coverage}~(b) and  (c), we transfer the knowledge from CLIP-ViT-L/14 to our Distill-UniCL* and compare with UniCL*. Both Distill-UniCL* and UniCL* are trained on the same relatively-small public datasets besides UniCL* requires the paired image-text data. In Fig.~\ref{fig:conceptual_coverage}~(b), we find distillation from the huge teacher model overall performs better than contrastive pretraining when there are a few conceptually-related training examples. By adding a significant amount ($>1$ million) of conceptually-related images (see the comparison of between Fig.~\ref{fig:conceptual_coverage}~(c) and (b) on ImageNet and Oxford-Flowers 
 datasets), the contrastive pretraining gets closer zero-shot performance to our distillation method but the contrastive pretraining requires the additional pairing information from the human annotators. For those datasets which only get fewer than 0.25 million of new images (such as Caltech-101 and EuroSAT datasets), the vision-language distillation still preserves the superior performance to the contrastive pretraining. 
\end{itemize}

%% file: A2_mmd_score_among_corpora.tex
\section{MMD among Image and Text Corpora}~\label{sec:mmd}
\begin{figure}
\begin{center}
     \includegraphics[width=0.5\linewidth]{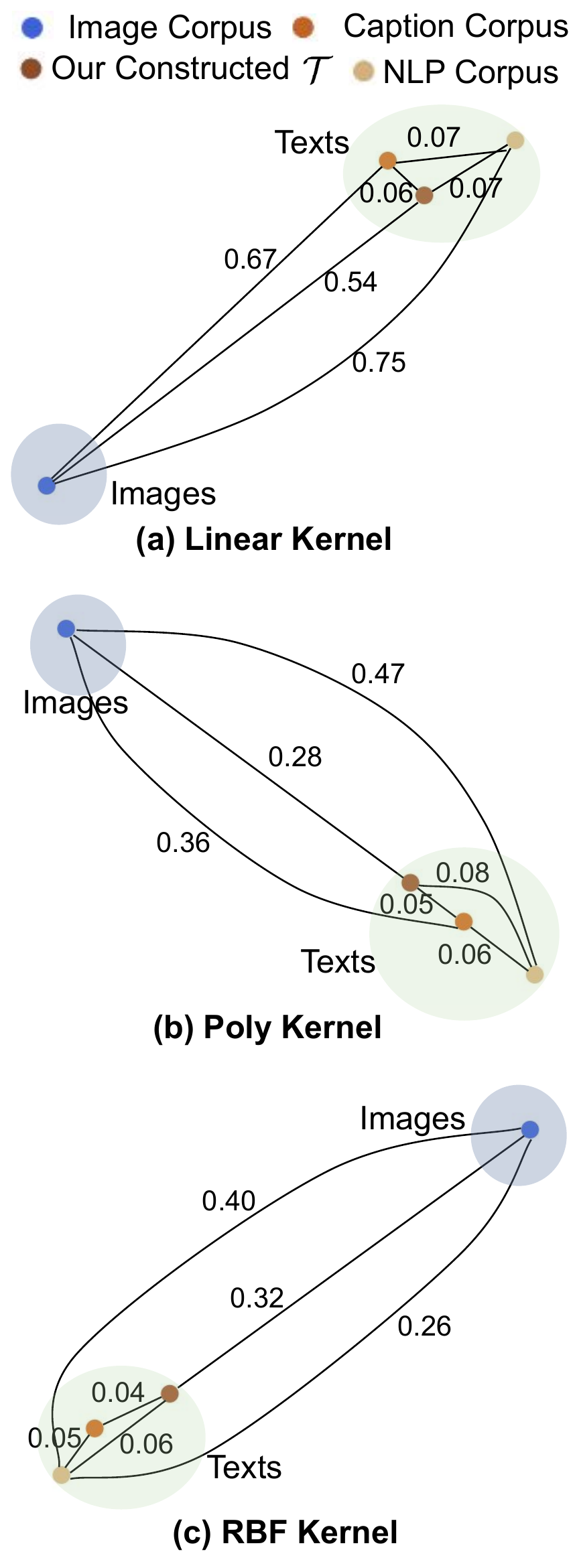}
\end{center} 
   \vspace{-20pt}
   \caption{\small 
   \textbf{MMD among four different image/text corpora's in the shared feature space.} We put MMD value on each edge.}
   \label{fig:mmd_distance} 
      \vspace{-15pt}
\end{figure}

We compute Maximum Mean Discrepancy (MMD) with of four image/text corpora's distributions in CLIP-ViT-L/14's shared feature space using linear, polynomial and RBF kernel respectfully. We use multi-dimensional scaling (MDS) to plot relatively locations of four corpora in a 2D space (see Fig.~\ref{fig:mmd_distance}). We have some observations of MMD analysis: 
\begin{itemize}[leftmargin=*]
    \item There exists clear modality gap in the shared feature spaces between the image and text features. It consists with the T-SNE plot in Fig.~5 (in the main paper).
    \item With Algorithm 1, our constructed text corpus is closer to visually-grounded caption corpus than the NLP corpus. It consists with the T-SNE plot in Fig.~4 (in the main paper).
    \item In the main paper, we show the our selected sentence is closer to the query image that its human-annotated captions. Here, we further show our constructed text corpus is also closer to the image corpus in the distribution level.
\end{itemize}

%% file: A8_full_ablation_study.tex
\begin{figure*}
\begin{center}
     \includegraphics[width=\linewidth]{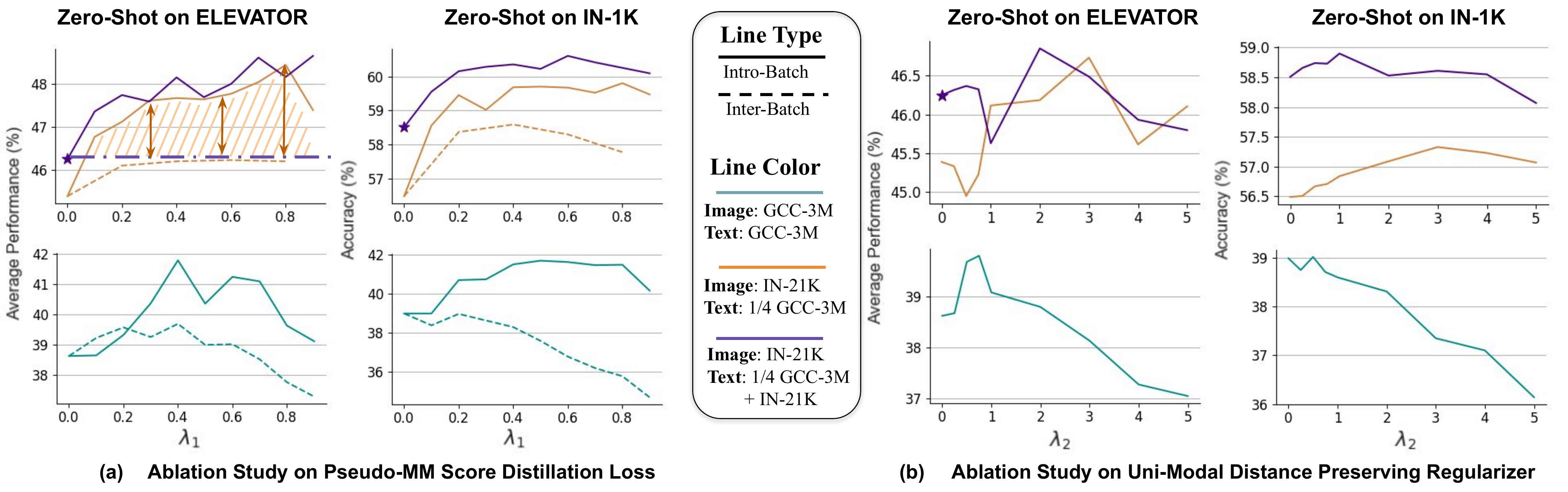}
\end{center} 
\vspace{-20pt}
    \caption{\small 
   \textbf{Ablation Studies on Pseudo-VL Score Distillation Loss $\mathcal{L}_{p\text{-}vl}$ and Uni-Modal Distance Preserving Regularizer $\mathcal{L}_{udist}$.} When using IN-21K images as image corpus in both ablation studies, using $\Lcal_{p\text{-}vl}$ and $\Lcal_{udist}$ better utilizes image embeddings. Sometimes, it even performs better than directly incorporating class names in  $\Lcal_{vl}$ (i.e the {\color{peru} Orange} Curve is sometimes above the {\color{indigo} Purple} \enquote{-- $\cdot$ -- $\cdot$} line. )}    
   \label{fig:loss_ablation} 

\end{figure*}

\input{tables/detailed_performance.tex}

\section{Full Ablation Studies on Losses}~\label{sec:full_ablation_on_losses}
In addition to Sec. 4.5 in the main paper, we provide our full ablation study on losses with different image and text training data. We show the effectiveness of $\Lcal_{p\text{-}vl}$ and $\Lcal_{udist}$ by Zero-Shot on ELEVATER and IN-1K in three realistic settings (see Fig.~\ref{fig:loss_ablation}): (1). The image concepts are entirely overlapped with visually-ground sentences ({\color{darkcyan} `Cyan'} curves). (2). The image concepts and the visually-grounded sentences are independent ({\color{peru} `Orange'}). (3). The image concepts are partially covered by the visually-grounded sentences  ({\color{indigo} `Purple'}).

\noindent \textbf{Ablation on $\mathcal{L}_{p\text{-}vl}$.} We gradually put more weights on the $\mathcal{L}_{p\text{-}vl}$  by increasing $\lambda_1$ in Eq.6 (main paper) from 0 to 1 in Fig.~\ref{fig:loss_ablation}~(a).  We compare the inter-batch version (i.e. $\uv_i$ and $\uv_j$ in Eq.10 from different batches) and intro-batch version (i.e. $\uv_i$ and $\uv_j$ from the same batch) of $\Lcal_{p\text{-}vl}$ and find the intro-batch $\Lcal_{p\text{-}vl}$ performs better than inter-batch $\Lcal_{p\text{-}vl}$ in setting (1) and (2), so we keep intro-batch version in other experiments. Furthermore, adding $\Lcal_{p\text{-}vl}$ with $\lambda_1 \le 0.9$ brings  better zero-shot performance than only using $\Lcal_{vl}$ in all three settings. However, we observe the dramatic performance drop when we totally replace $\Lcal_{vl}$ with $\Lcal_{p\text{-}vl}$ (\ie $\lambda_1=1$). We argue improvement with smaller $\lambda_1$'s and drop at $\lambda_1 =1$ both due to the gap between images and text embeddings in the shared feature space.  
Interestingly, we find that the performance with $\lambda_1 > 0$ in setting (2) is better than using pure $\Lcal_{vl}$ ($\lambda_1 = 0$) in setting (3).
It indicates that $\Lcal_{p\text{-}vl}$ is more effective than prompt sentences of class names since using image embeddings as pseudo text embeddings in $\Lcal_{p\text{-}vl}$ introduces richer concepts than class names.

\noindent \textbf{Ablation on $\mathcal{L}_{udist}$.} We increase $\lambda_2$ from 0 to 5, to introduce $\Lcal_{udist}$ as a regularization term. Generally, $\Lcal_{udist}$ benefits the Zero-Shot on ELEVATER since it tries to preserve the geometry of image features. But different $\lambda_2$ works the best for different datasets. $\Lcal_{udist}$ slightly improves IN-1K performance when $\lambda_2$ is small but it quickly harms IN-1K performance when $\lambda_2$ gets larger. We suspect the poor student embedding in early training along with the large regularization term detours the gradient decent trajectory. $\Lcal_{udist}$ is more effective when the text does not cover the image concepts in setting (2), where it still improves Zero-Shot on ELEVATER and IN-1K with a large $\lambda_2$. Our main experiment (Table.~1 in the main paper) further shows $\Lcal_{udist}$ is less effective when applying $\Lcal_{p\text{-}vl}$ and $\Lcal_{udist}$ together. 

%% file: tables/detailed_performance.tex
\begin{table*}
    \begin{center}
        \resizebox{0.9\linewidth}{!}{
        \begin{tabular}{c | c c | c c c| c c c}
            \Xhline{3\arrayrulewidth} 
            \multirow{2}{*}{Downstream Tasks} &  \multicolumn{2}{c|}{ViT-B/32} &  \multicolumn{3}{c|}{Swin-Tiny (IN-21K)} & \multicolumn{3}{c}{Swin-Tiny (IN-21K+YFCC-14M) } \\ 
            & \# CR-Images & \ours & \# CR-Images & UniCL* & \ours & \# CR-Images & UniCL* & \ours  \\
            \Xhline{3\arrayrulewidth} 
            Hateful Memes & 3.1K & 54.4\% & 80 & 53.9\% & 53.1\%  & 322 & 52.9\% & 53.6\%  \\
            PatchCamelyon & 158 & 52.9\% & 0 & 49.6\% & 52.9\% & 15 & 51.5\% & 50.4\% \\
            Rendered-SST2 & 3.4 K & 50.0\% & 650 & 49.9\% & 9.4\% & 3.2K & 49.8\% & 49.8\% \\
            KITTI Distance  & - & 9.28\% & - & 24.6\%  & 22.2\% & - & 12.4\% & 10.3\% \\
            FER 2013 & 579 & 39.5\% & 432 & 24.0\% & 92.2\% & 467 &25.3\% & 27.4\%  \\
            CIFAR-10 & 0.6M & 95.5\% & 5.9K & 91.2\% & 92.2\% & 335.8K & 89.3\% & 90.2\% \\
            EuroSAT & 68.0K & 54.0\% & 0 & 27.2\%  & 30.4\% & 46.4K & 36.5\% & 46.1\% \\
            MNIST &  1.1K & 38.8\% & 0 & 11.64\% & 21.6\% & 619 & 13.0\% & 20.1\%\\
            VOC 2007 Classification & 1.1M & 83.4\% & 3.3K & 82.0\% & 83.3\% & 544.4K & 82.9\% & 83.4\% \\
            Oxford-IIIT Pets & 9.1K & 88.6\% & 3.3K & 47.6\% & 70.1\% & 5.6K & 69.1\% & 74.3\%  \\
            GTSRB &  610 & 27.6\% & 0 & 7.7\% & 16.2\% & 545 & 11.8\% & 15.0\%\\
            Resisc-45  & 1.3M & 55.2\% & 7.8K & 21.6\% & 29.3\% & 955.2K & 32.3\% & 40.3\% \\
            Describable Textures & 5.2K & 52.6\% & 5.2K & 37.7\% & 46.38\% & 4.4K & 42.0\% & 47.4\% \\
            CIFAR-100 & 3.8M & 76.4\% & 42.2K & 66.8\% & 71.0\%  & 1.9M & 64.1\% & 68.9\%\\
            FGVC Aircraft (variants) & 90 & 20.1\% & 0 & 3.0\% & 6.8\%  & 0 & 9.1\% & 12.8\%\\
            Food-101 & 57.7K & 80.4\% & 13.8K & 57.0\%  & 69.3\% & 41.9K & 65.9\% & 71.7\% \\
            Caltech-101 & 1.0M & 89.6\% & 28.6K & 82.5\% & 85.2\% & 475.4\% & 84.4\% & 85.7\%  \\
            Oxford Flowers 102 & 2.9M & 75.9\% & 26.7K & 73.4\% & 79.0\% & 2.9M & 77.9\% & 77.0\% \\
            Stanford Cars & 0 & 43.35\% & 0 & 2.3\%  & 5.0\% & 0 & 8.0\% & 12.5\%  \\
            Country-211 & - & 17.1\%  & - & 3.4\%  & 7.5\% &- & 13.5\% & 14.9\% \\
            Imagenet & 2.1M & 60.8\% & 0 & 51.4\% & 59.5\% & 1.3M & 58.7\% & 60.0\% \\
             \Xhline{3\arrayrulewidth} 
        \end{tabular}
        } 
        \caption{\small \textbf{Conceptually-Related Images and Different Models' Zero-Shot Performance on Each Dataset}. We provide the zero-shot performance of every model proposed in this paper for each dataset. 
        We also report the number of conceptually-related images (CR-Images).}~\label{table:detailed-performance}
    \end{center}
    \vspace{-10pt}
\end{table*}

%% file: A7_detailed_information.tex
\section{Detailed Performance on Each Dataset}~\label{sec:detailed_performance}
In this section, we  provide the zero-shot performance of models proposed in this paper for each downstream dataset.  Other public models' (such as CLIP~\cite{radford2021learning} or UniCL~\cite{yang2022unified}) performance can computed by ELEVATER toolkit~\cite{li2022elevater} (see Table.~\ref{table:detailed-performance}).   We also report the number of conceptually-related images computed from Table 11 of ~\cite{yang2022unified} for each downstream dataset. Furthermore, we report the detailed performance on five datasets with the domain shift to ImageNet-1K (see Table~\ref{table:robustness}) as the supplement to Table 1 in the main paper.

\newpage

\input{tables/robustness}

\clearpage

%% file: tables/robustness.tex
\begin{table*}
    \begin{center}
        \resizebox{0.9 \linewidth}{!}{
        \begin{tabular}{c c c c c c c }
            \Xhline{3\arrayrulewidth} 
            Models & ImageNet-v2 & ImageNet-R & ObjectNet & ImageNet-Sketch & ImageNet-A & \cellcolor{yellow!15}  Average\\
            \Xhline{3\arrayrulewidth}
            CLIP-ViT-B/32 & 55.9\% & 69.0\% & \textbf{44.2\%} & 42.3\% & 31.5\% & \cellcolor{yellow!15}  48.6\% \\
            Distill-ViT-B/32 (Captions) & 57.5\% & 69.7\% & 42.5\% & 45.7\% & 31.6\% & \cellcolor{yellow!15}  49.4\% \\
            Distill-ViT-B/32 (NLP Texts) & \textbf{58.9\%} & \textbf{69.8\%} & 43.2\% & \textbf{46.5\% }& \textbf{32.2\%} & \cellcolor{yellow!15}  \textbf{50.2\% }\\
             \Xhline{3\arrayrulewidth} 
        \end{tabular}
        } 
        \caption{\small \textbf{Robustness.} Our Distilled-ViT-B/32 models perform better than CLIP-ViT-B/32 model on 5 datasets which have distribution shift to origin ImageNet-1K data. The results demonstrates our distilled models preserve the robustness to the distribution shift.} \label{table:robustness}
    \end{center}
\vspace{-35pt}
\end{table*}